\pdfoutput=1
\documentclass{bmvc2k}

\title{TeCoNeRV: Leveraging Temporal Coherence for Compressible Neural Representations for Videos}

\addauthor{Namitha Padmanabhan}{namithap@umd.edu}{1}
\addauthor{Matthew Gwilliam}{mgwillia@umd.edu}{1}
\addauthor{Abhinav Shrivastava}{abhinav@cs.umd.edu}{1}

\addinstitution{
 University of Maryland\\
 College Park, Maryland, USA\\
}

\runninghead{Padmanabhan, Gwilliam, Shrivastava}{TeCoNeRV}

\usepackage{microtype}
\usepackage{graphicx}
\usepackage{dblfloatfix}
\usepackage{subcaption}
\AtBeginDocument{%
  \setlength{\abovecaptionskip}{5pt}%
  \setlength{\belowcaptionskip}{5pt}%
}
\usepackage{booktabs}
\usepackage{placeins}

\usepackage{amsmath}
\usepackage{amssymb}
\usepackage{mathtools}
\usepackage{amsthm}

\usepackage[capitalize,noabbrev]{cleveref}
\Crefname{appendix}{Appendix}{Appendices}

\usepackage{makecell}
\usepackage{colortbl}
\usepackage{multirow}
\usepackage{contour}
\usepackage{threeparttable}
\usepackage{pifont}
\usepackage[dvipsnames]{xcolor}
\usepackage{adjustbox}

\newcommand{\cmark}{\checkmark}
\newcommand{\xmark}{\ding{55}}

\input{suppl_style}

\begin{document}

\maketitle

\begin{abstract}
\label{sec:abstract}
    Implicit Neural Representations (INRs) have recently demonstrated impressive performance for video compression.
    However, since a separate INR must be overfit for each video, scaling to high-resolution videos while maintaining encoding efficiency remains a significant challenge.
    Hypernetwork-based approaches predict INR weights (hyponetworks) for unseen videos at high speeds, but with low quality, large compressed size, and prohibitive memory needs at higher resolutions. 
    We address these fundamental limitations through three key contributions: 
    (1) an approach that decomposes the weight prediction task spatially and temporally, by breaking short video segments into patch tubelets, to reduce pretraining memory overhead across resolutions; (2) a residual-based storage scheme that captures only differences between consecutive segment representations, significantly reducing bitstream size; and (3) a temporal coherence regularization framework that encourages changes in the weight space to be correlated with changes in video, making residuals even smaller.
    Our proposed method, TeCoNeRV, achieves substantial improvements of 2.47dB and 4.94dB PSNR over the baseline at 480p and 720p on UVG, with 36\% lower bitrates and 1.5-3${\times}$ faster encoding speeds.
    With our low memory usage, we are the first hypernetwork approach to demonstrate results at $480$p, $720$p and $1080$p on UVG, HEVC and MCL-JCV.
    Our project page is available \href{https://namithap10.github.io/teconerv/}{here}.
\end{abstract}

\section{Introduction}
\label{sec:intro}

\begin{figure}[t]
    \vspace{0.5em}
    \centering
    \includegraphics[width=\linewidth]{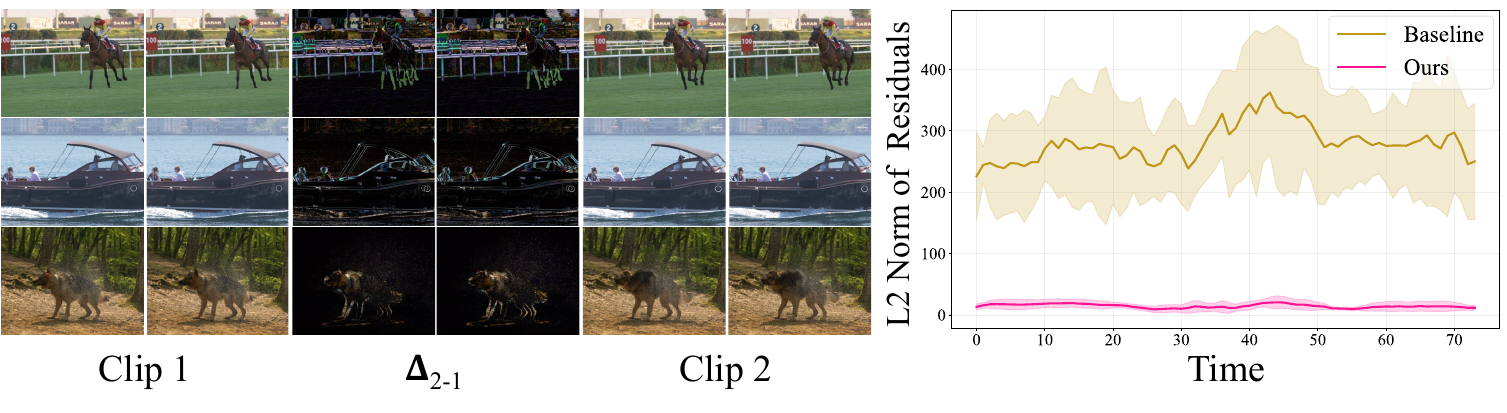}
    \caption{
    \textbf{TeCoNeRV achieves smoother weight transitions across video clips, enabling superior compression.} Left: Selected frames from consecutive clips of three videos from the UVG dataset, with the change in pixel space visualized in the middle column. Right: L2 norm of clip-to-clip weight residuals over time for the baseline (NeRV-Enc~\cite{chen2024fast}) versus our approach (TeCoNeRV). Our temporal coherence regularization produces smaller weight residuals as video content evolves, resulting in efficient compression while preserving visual quality.}
    \label{fig:teaser}
\end{figure}

Video data continues to dominate internet traffic, creating an ever-growing need for efficient compression. 
Conventional block-based video codecs~\cite{le1991mpeg,wiegand2003overview,sullivan2012overview,bross2021overview} establish strong compression-quality tradeoffs through decades of engineering, while neural  compression methods~\cite{rippel2019learned, khani2021efficient, li2021deep, li2023neural} offer quality and size improvements, although at the cost of slower decoding.
Recently, Implicit Neural Representations (INRs) have shown promise by representing videos as compact neural networks~\cite{chen2021nerv, kwan2023hinerv, kim2024c3}, achieving fast decoding speeds through simple feed-forward operations. 
However, traditional INRs train a separate neural network for each video, with encoding typically taking several hours per sequence, thus making it prohibitively slow for practical applications.
To address the problem of INR encoding efficiency, hypernetwork methods train a single network to predict the weights of the individual INRs (called ``hyponetworks'') for unseen videos~\cite{chen2022transformers,kim2022generalizable,chen2024fast}, reducing encoding time by multiple orders of magnitude. %
Nevertheless, prior hypernetwork compression works demonstrate results only at low resolutions ($256{\times}256$), for sparse framerates (8 to 16 frames per video), and with large compressed sizes.

We propose TeCoNeRV, a framework for adapting INR hypernetworks to compress videos efficiently at higher resolutions.
Using prior works, predicting hyponetwork weights becomes computationally prohibitive, since memory requirements scale quadratically as resolution increases.
We address this fundamental scaling limitation by introducing a patch-tubelet approach that decomposes each clip into smaller spatio-temporal volumes. 
This decomposition (i) alleviates the memory requirements for predicting large hyponetworks, and (ii) directly lends itself to inference at higher frame resolutions.
The latter is possible even when the patch size is not a perfect factor of the frame size because we can employ strided patches with overlap.
We simply crop at the overlapping patch boundaries to ensure smooth transitions.
This unlocks a huge benefit: resolution-independent training. 
For example, we can train our model on $480$p videos and perform inference at $1080$p.

Further, we analyze the implicit weight representations produced by existing hypernetwork systems for videos. 
We find that na\"ive hypernetwork predictions generate weights that can vary dramatically for adjacent frames (\Cref{fig:teaser}, right panel), even when visual changes are minor (left panel).
We argue that smooth pixel transitions should correspond to smooth weight transitions for better compression.
Therefore, we introduce a temporal coherence regularization that implicitly conditions weight spaces to evolve smoothly across adjacent clips.
Applied as finetuning after initial training, this naturally induces sparsity in weight transitions.
This regularization dramatically reduces both the magnitude and variance of weight residuals across clips, as evidenced by the lower and more consistent L2 norm curve of our approach in \Cref{fig:teaser}, compared to the baseline (NeRV-Enc~\cite{chen2024fast}). 
We store only the first clip's full weights and compact residuals for subsequent clips, significantly reducing bitstream size without sacrificing quality.
The strength of the regularization can thus serve as a rate control mechanism.

At inference time, our method processes the target video clip-by-clip, predicting hyponetwork weights for each patch position and storing only the first prediction along with subsequent compact residuals. 
This approach maintains encoding speeds comparable to NeRV-Enc~\cite{chen2024fast} while achieving superior compression and reconstruction quality. 
TeCoNeRV achieves substantial improvements over the baseline on the UVG dataset~\cite{mercat2020uvg}: 2.47dB PSNR gain at 480p and 4.94dB gain at 720p, with a 36\% reduction in bitrate, while maintaining practical encoding speeds.
We also show consistent gains on the diverse Kinetics-400~\cite{kay2017kinetics}, HEVC~\cite{ohm2012comparison} and MCL-JCV~\cite{wang2016mcl} datasets.
In addition, encoding residuals over individual clip weights significantly improves encoding speed. We summarize our main contributions below:
\begin{itemize}
    \item TeCoNeRV, the first hypernetwork-based approach that successfully scales to higher resolution video compression by addressing fundamental limitations in memory usage and encoding efficiency. We achieve this through a patch-tubelet decomposition strategy that divides the weight prediction task spatially and temporally.
    \item A residual encoding strategy that dramatically reduces bitstream size by exploiting temporal redundancy between consecutive clips.
    \item Rate control with weight-space conditioning that aligns hyponetwork evolution with video content, ensuring smoother transitions that produce smaller, more consistent residuals across clips for further compression.
    \item Resolution-independent training -- we can predict high resolution videos using models trained at any resolution. 
\end{itemize}

\FloatBarrier
\section{Related Work}
\label{sec:background}

\noindent\textbf{Video Compression.}
Traditional video codecs~\cite{le1991mpeg,wiegand2003overview,sullivan2012overview} employ block-based motion estimation and compensation techniques to achieve acceptable compression-quality tradeoffs.
With the rise of deep learning, many works have sought to improve or replace components of these pipelines~\cite{li2021deep, rippel2019learned, khani2021efficient, li2023neural, li2024neural}. 
The NVC and DCVC family~\cite{li2021deep,li2023neural,li2024neural,Jia_2025_CVPR,Li_2026_CVPR} achieves competitive rate-distortion performance through techniques like contextual entropy modeling and temporal feature modulation.
Through years of operational cost optimization, DCVC-RT~\cite{Jia_2025_CVPR} achieves real-time performance for 1080p encoding.
While these methods excel in compression ratio, INR-based approaches offer complementary advantages including fast decoding without complex motion modules, compact storage, and support for continuous spatiotemporal interpolation.

\noindent\textbf{Implicit Neural Representations.}
INRs encode signals such as images, videos, or 3D scenes as neural networks that map continuous coordinates to signal values~\cite{mildenhall2020nerf,sitzmann2020implicit}.
For images, they map spatial coordinates $(x,y)$ to RGB values~\cite{tancik2020fourfeat,dupont2021coin,dupont2022coin++,ladune2023cool}, while for videos, they incorporate temporal coordinates ($t$)~\cite{zhang2021implicit, kim2024c3,leguay2024cool}, offering compact storage suitable for compression tasks. 
NeRV~\cite{chen2021nerv} maps time embeddings to entire frames, spawning a family of video INRs~\cite{chen2023hnerv,kwan2023hinerv,lee2023ffnerv,He_2023_CVPR,li2022nerv,yan2024ds,zhao2023dnerv,kwan2024nvrc,gupta2024pnerv} advancing different aspects of frame-wise representation. 
Recent works model flow in standard INRs~\cite{lee2023ffnerv,shin2024efficient} or optimize spatio-temporal patches directly for static signals~\cite{saragadam2022miner,ashkenazi2024towards}. 
In contrast, our method operates in the generalizable hypernetwork paradigm for videos, without explicit flow modeling, and leverages temporal coherence to encourage small residuals between similar frame groups. 
ActINR~\cite{Kayabasi_2025_CVPR} shares INR weights across a video's frames while modulating per-frame biases, a parameterization strategy distinct from hypernetworks that predict weight modulation for unseen video.
NIRVANA~\cite{maiya2023nirvana} addresses encoding efficiency by encoding frame group patches autoregressively, however it still requires per-video optimization, taking hours to encode. 
Our hypernetwork approach eliminates per-video optimization entirely.

\noindent\textbf{Hypernetworks.}
To address the per-sample long encoding times that INRs suffer from, several works have explored hypernetworks -- neural networks that predict the weights of another network (the hyponetwork). 
Some works demonstrate hypernetwork approaches for 3D scenes~\cite{sitzmann2020scene, chiang2022stylizing, sen2022inr,sen2023hyp} while others focus on predicting INRs to generate images~\cite{skorokhodov2021adversarial} or videos~\cite{yu2022generating,ren2025nerv}. 
Recent work focuses on hypernetworks predicting INR weights to reconstruct inputs~\cite{chen2022transformers,kim2022generalizable,maiya2024latent}. 
NeRV-Enc/NeRV-Dec~\cite{chen2024fast} introduces a single hypernetwork for frame-wise video representation with faster encoding speeds than pixel-wise approaches. 
However, existing hypernetwork compression methods struggle with low quality and practical resolutions and video lengths.
We address these scalability challenges through patch-tubelet decomposition, combined with residual encoding and temporal coherence regularization to improve compression efficiency.

\noindent\textbf{Feature Representation Learning.} 
Our temporal coherence approach draws inspiration from self-supervised learning techniques~\cite{chen2020simple,he2020momentum,grill2020bootstrap,sun2021composable} that exploit temporal structure in videos. \citet{jayaraman2016slow} employ unsupervised contrastive loss to learn representations that change smoothly with visual transformations, extending earlier slow feature analysis work~\cite{wiskott2002slow, hurri2003simple}. Other methods use time as a semantic similarity proxy~\cite{sermanet2018time}, predict future latents~\cite{oord2018representation}, or exploit temporal order as supervision~\cite{wang2016actions,Lee_2017_ICCV,misra2016shuffle}. These insights motivate our approach -- conditioning weights to evolve synchronously with temporal progression exploits video continuity for compression efficiency.

\section{Method}
\label{sec:method}

\subsection{Background and Problem Formulation}
\label{subsec:problem}
Given an input video $V\in\mathbb{R}^{T{\times}3{\times}H{\times}W}$ with $T$ frames at spatial resolution $H{\times}W$, an INR represents the video as a neural network $f_\theta$ mapping spatiotemporal coordinates to RGB values. 
We build upon NeRV-Enc/NeRV-Dec~\cite{chen2024fast}, which uses a hypernetwork $g_\phi$ to predict weights of a NeRV~\cite{chen2021nerv} video representation (HypoNeRV).
Unlike NeRV-Enc's sparse frame sampling, we process consecutive frame sequences (clips): $S \in \mathbb{R}^{N{\times}3{\times}H{\times}W}$, where $N$ is the clip length. 
For each clip, $g_\phi$ predicts weights of the hyponetwork $f_\theta$ to reconstruct frames in the sequence. %
The hyponetwork follows the NeRV architecture with time positional encoding and NeRV blocks (convolution $+$ PixelShuffle~\cite{shi2016real}).
The learnable parameters comprise: (1) video-agnostic base parameters $\theta_{\text{base}}$, shared across all videos, and (2) clip-specific unique parameters $\theta_{\text{uniq}}$, generated by the hypernetwork encoder from input data tokens and learnable weight tokens.
The final hyponetwork weights $\theta$ are obtained by modulating the base parameters with expanded unique parameters, through element-wise multiplication. 
This system is optimized via mean squared error between reconstructed and ground truth frames:
\begin{equation}
\begin{aligned}
\label{eqn:mse_opt}
    (\phi^*, \theta_{\text{base}}^*) =
    \mathop{\arg\!\min}\limits_{\phi,\, \theta_{\text{base}},\, \theta_{\text{uniq}}}
    \sum_{V \in D_{\text{train}}} \sum_S \sum_t
    \|f_{\theta = g_\phi(S)}(t) - S_t\|_2^2
\end{aligned}
\end{equation}
where $D_{\text{train}}$ is the training set, $S$ are clips from video $V$, $t$ indexes frames within each clip. 

In the following sections, we explain the main components of our proposed method -- a training paradigm for high-resolution videos, differential encoding of clip weights using residuals and finally, the weight space regularization framework for temporal coherence.

\begin{figure}[t]
    \centering
    \includegraphics[width=\linewidth]{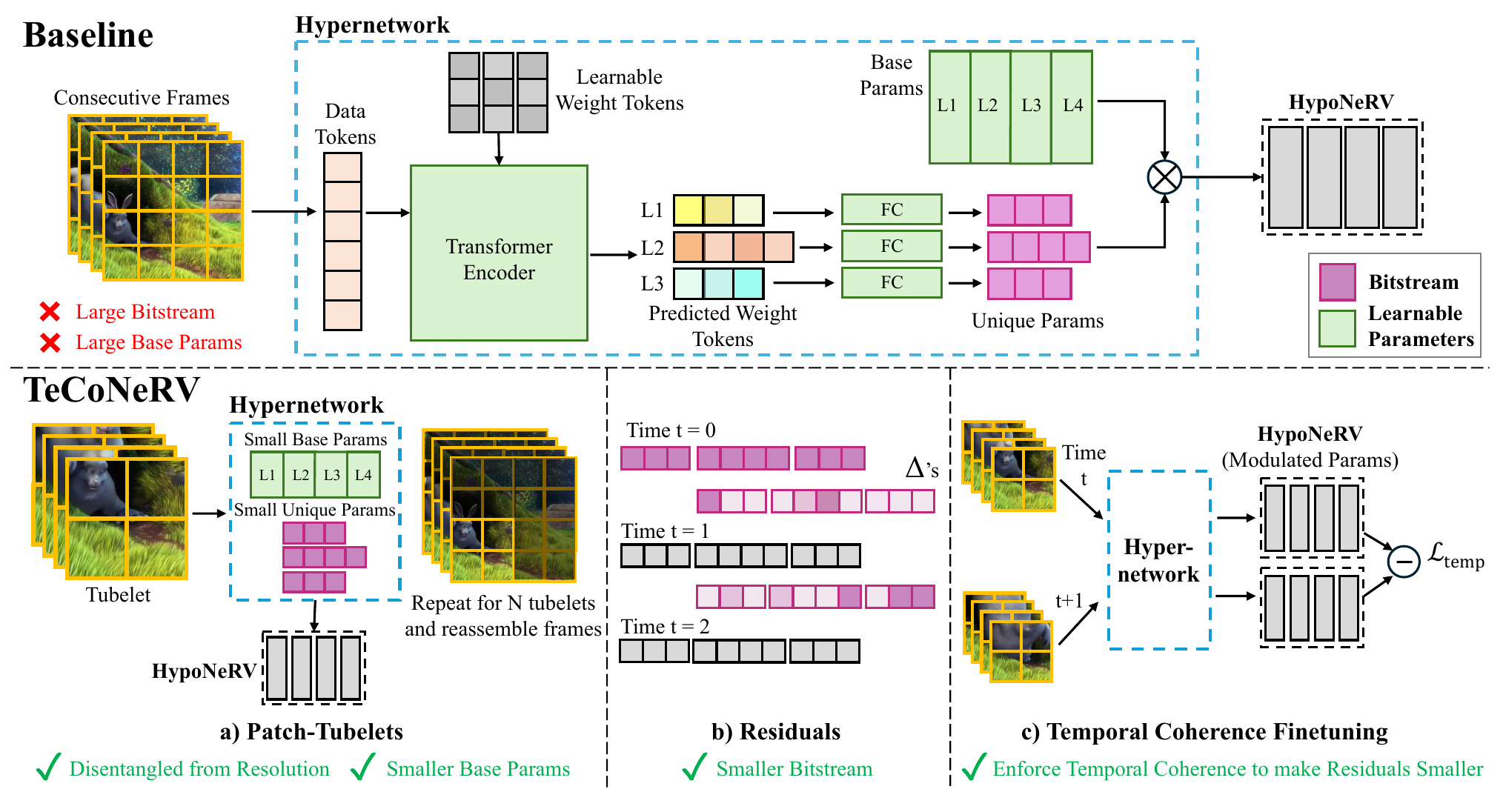}
    \caption{\textbf{Overview of TeCoNeRV.} Above: Hypernetworks from prior work predict weights for entire video frames at once, resulting in large base parameters and bitstream size. Below: Our approach with (a) patch-tubelets that decouple spatial resolution from memory requirements, (b) residual encoding that stores only weight differences across time steps, and (c) temporal coherence finetuning that regularizes weight differences. Together, these components achieve better compression efficiency and superior reconstruction quality. 
    }
    \label{fig:method}
\end{figure}

\subsection{Scaling to Higher Resolution Videos}
\label{subsec:fullres}
A key challenge with prior hypernetwork-based approaches is that memory requirements scale quadratically with video resolution, making direct application to higher-resolution videos ($720$p+) infeasible. To address this limitation, we introduce patch-tubelets (\Cref{fig:method}), which enable our approach to scale efficiently to higher resolutions.
We partition each clip $S \in \mathbb{R}^{N{\times}3{\times}H{\times}W}$ into patch tubelets $P \in \mathbb{R}^{N{\times}3{\times}H_p{\times}W_p}$, where $H_p{\ll}H$ and $W_p{\ll}W$ represent the spatial dimensions of each patch. This decomposition reduces the problem from predicting weights for entire frames to localized spatio-temporal volumes, making the weight prediction problem more tractable.
This approach offers two advantages: memory efficiency, as requirements depend on patch dimensions rather than full-frame dimensions; and resolution independence, as the hypernetwork processes fixed-size patches regardless of the full video resolution.
At inference time, we predict hyponetwork weights for each spatial position and combine their outputs to reconstruct full-resolution frames. 
Quality metrics are computed at native resolution to ensure fidelity to the original signal.
This methodology is inherently scalable to higher resolutions, as computational complexity grows linearly with the number of patches rather than quadratically with resolution.

\subsection{Residuals: Compression and Storage}
\label{subsec:residual}
We improve compression efficiency by exploiting temporal redundancy across consecutive clips. 
Instead of storing complete unique parameters for each clip, we implement a differential encoding scheme where for each spatial position (patch tubelet) in the video, we store complete unique parameters $\theta_{\text{uniq}}$ only for the first clip, and compact residuals for subsequent clips.
During decoding, clip parameters are reconstructed via cumulative addition of previous-clip residuals, before modulating with the shared base parameters installed as part of the decoder.
This approach is particularly effective for consecutive frames with similar content, where weight transitions are small. The compression becomes even more effective when combined with our temporal coherence regularization described next.
For the final compressed bitstream, we apply uniform quantization to the first hyponetwork weights and residuals followed by arithmetic coding \cite{witten1987arithmetic}, without requiring quantization-aware training.

\subsection{Temporal Coherence Regularization}
\label{sec:temp_coh}

While residual encoding exploits redundancy, hypernetworks trained solely for reconstruction do not inherently produce weight spaces that transition smoothly over time.
This phenomenon is also observed in per-video INRs optimized with pixel-level losses; prior analysis reveals that individual neurons do not maintain stable correspondence to semantic entities over time~\cite{Padmanabhan_2024_CVPR}.
Thus, even visually smooth transitions can induce abrupt changes in the predicted hyponetwork weight space. %
To address this, we introduce a temporal coherence regularization that implicitly conditions hyponetwork weights to evolve smoothly with video content.
This is applied as a finetuning step after training, with the combined objective:
\begin{equation}
\label{eqn:temp_finetune}
    (\phi^*, \theta_{\text{base}}^*) = \mathop{\arg\!\min}\limits_{\phi,\, \theta_{\text{base}},\, \theta_{\text{uniq}}} \mathcal{L}_{\text{recon}}(\phi, \theta) + \lambda_{\text{temp}} \mathcal{L}_{\text{temp}}(\theta)
\end{equation}
where $ \mathcal{L}_{\text{recon}}(\phi, \theta)$ is the MSE loss from \Cref{eqn:mse_opt}, $\mathcal{L}_{\text{temp}}$ is our unsupervised temporal regularization loss, and $\lambda_{\text{temp}}$ is the hyperparameter that controls the regularization strength. To implement this loss, we construct a dataset of consecutive patch-tubelet pairs, $D_{\text{reg}}$~\cite{jayaraman2016slow}:
\begin{equation}
    D_{\text{reg}} = \left\{ (i, i+1) : P_i, P_{i+1} \text{ are consecutive tubelets in } V \in D_{\text{train}} \right\}
\end{equation}
The regularization loss penalizes first-order differences between consecutive weight spaces:
\begin{equation}
\begin{aligned}
\label{eqn:temp_loss}
        \mathcal{L}_{\text{temp}} =  \sum_{(i,i+1) \in D_{\text{reg}}} | \theta_{i} - \theta_{i+1} |
\end{aligned}
\end{equation}
where $\theta_i = g_\phi(S_i)$ are the hyponetwork weights for clip $i$, and $| \theta_{i} - \theta_{i+1} |$ is the $\ell_1$-norm distance metric.
Although this loss is not explicitly content-weighted, penalizing the first-order differences jointly with reconstruction selects smooth trajectories among near-equivalent weight-space solutions, thus yielding more compressible residuals.
We apply this regularization to the modulated parameters $\theta$ rather than the unique parameters $\theta_{\text{uniq}}$, as we find this yields better rate-distortion performance.
This parallels findings in contrastive learning, where losses applied to projected representations are often more effective than those applied to stored features~\cite{chen2020simple}.

\FloatBarrier
\section{Experimental Setup}
\label{sec:expt_setup}

\subsection{Dataset and Training Details}
\label{sub:setup}
We train on Kinetics-400~\cite{kay2017kinetics}, which contains 240k training videos and 20k test videos in 400 classes, each approximately 10 seconds long.
Following~\cite{chen2024fast}, we use a subset of 10,000 videos with 25 videos per class for training.
We evaluate at three resolutions -- $480$p ($640{\times}480$), $720$p ($1280{\times}720$) and $1080$p ($1920{\times}1080$) -- and report peak signal-to-noise ratio (PSNR), structural similarity index measure~\cite{wang2004image} (SSIM) and bits-per-pixel (bpp) with encoding and decoding frames-per-second (FPS).
We evaluate on UVG \cite{mercat2020uvg}, a Kinetics-400 validation subset, MCL-JCV \cite{wang2016mcl}, and HEVC Classes B, C and E \cite{ohm2012comparison, flynn16common} which comprise videos at $1080$p, $480$p and $720$p respectively.
As memory requirements for \cite{chen2024fast} scale prohibitively with resolution, we provide comparison at $480$p and $720$p, and compare to NeRV~\cite{chen2021nerv} and~HiNeRV~\cite{kwan2023hinerv} at all resolutions.
For preprocessing, videos are resized on the shorter side and center cropped.
During training, a random patch-tubelet position is selected, and 8 consecutive frames are sampled per video.
This contrasts with \citet{chen2024fast} which encodes 8 evenly sampled frames across the video. 
During inference, we predict hyponetwork weights for all patch positions in each 8-frame clip, storing first predictions and previous-clip residuals for subsequent clips.

\subsection{Baseline Details}
\label{sub:baselines}

Existing INR video compression methods can be categorized into two encoding regimes, distinguished by their encoding-time budgets.
Per-video optimization methods such as \cite{chen2021nerv,chen2023hnerv,kwan2023hinerv} overfit a neural representation to each input video, achieving strong reconstruction quality but requiring extended optimization, often several hours per sequence.
In contrast, hypernetwork methods such as NeRV-Enc~\cite{chen2024fast} predict INR weights in a single feed-forward pass, thus encoding a 1080p UVG video in seconds or minutes rather than hours.
Our method, TeCoNeRV, operates in the latter regime, focusing on fast encoding and scaling hypernetwork architectures to higher resolutions and longer videos. 
We therefore select NeRV-Enc as our primary baseline, as the only prior method explicitly designed for the fast-encoding regime.
Additionally, as memory requirements for NeRV-Enc scale prohibitively at high resolutions, we include NeRV and HiNeRV as representative per-video optimization baselines.

\textbf{NeRV-Enc* Model.}
We use the hypernetwork encoder from NeRV-Enc/NeRV-Dec~\cite{chen2024fast} with a 5-layer HypoNeRV and adapt settings for each target resolution.
Following insights from GINR \cite{kim2022generalizable} and NeRV-Enc, we allocate unique parameters with emphasis on the second layer.
The final layer utilizes only shared base parameters without modulation.
We refer to this clip-wise application of the baseline as NeRV-Enc*. 
Further details on hypernetwork and hyponetwork architectures are provided in \Cref{sec:supp_hypernet_details}.%

\textbf{NeRV and HiNeRV.}
We use the improved NeRV architecture from \citet{chen2023hnerv} and follow the rules given for scaling.
For HiNeRV~\cite{kwan2023hinerv}, we adapt the settings from the paper to the required resolution.
Allocating per-video methods their full optimization budget places their encoding time several orders of magnitude outside our target regime.
Thus, following our established practice~\cite{gwilliam2026design}, and for meaningful comparison across regimes, we configure NeRV and HiNeRV with encoding-time budgets (which include per-video training time) matching or exceeding our hypernetwork inference time and target bpp at least as high as our model.
Further details are provided in \Cref{sec:supp_nerv_hinerv_details}.%

\subsection{TeCoNeRV Details}
We adapt the NeRV-Enc* baseline for patch tubelets, and use a more compact 4-layer HypoNeRV while retaining the same hypernetwork encoder architecture.
We calibrate the hyponetwork parameter distributions to match the pre-quantization bpp of NeRV-Enc* for fair comparison.
The size of our base parameters is also much smaller than the baseline for the same train resolution (\Cref{tab:train_time_mem}).
For TeCoNeRV, we apply weight regularization finetuning after training the patch-based model (\Cref{eqn:temp_finetune}). 
For fair comparison, the baseline NeRV-Enc* and our patch-tubelet model also undergo 50 epochs of finetuning using only the reconstruction loss.
Through hyperparameter search, we determine an optimal value of $\lambda_{\text{temp}} = 0.1$, and apply regularization to the modulated parameters for stable rate-distortion performance.

\section{Results}
\label{sub:quant_res}

\begin{table*}
\caption{\textbf{Comparison of TeCoNeRV with baselines for video encoding at resolutions 480p, 720p and 1080p}. Results reported on UVG, HEVC Classes B, C and E, and MCL-JCV. Bits-per-pixel (bpp) is reported after quantization followed by arithmetic coding, and FPS denotes encoding / decoding speeds. Our method achieves superior PSNR and SSIM while using significantly fewer parameters. NeRV-Enc* refers to \citet{chen2024fast} adapted for high-resolution videos. TeCoNeRV\textsuperscript{1, 2, 3} indicate configurations of our model trained with patch sizes $320{\times}160$, $320{\times}240$ and $384{\times}270$ respectively.
}
\centering
\label{tab:allres_nonoverlap}
\setlength{\tabcolsep}{2pt}
\renewcommand{\arraystretch}{1.08}
\resizebox{\linewidth}{!}{
\begin{tabular}{@{}ll cc  ccc  ccc  ccc@{}}
\toprule
& & & & \multicolumn{3}{c}{UVG~\cite{mercat2020uvg}}
& \multicolumn{3}{c}{HEVC~\cite{ohm2012comparison}}
& \multicolumn{3}{c}{MCL-JCV~\cite{wang2016mcl}} \\
\cmidrule(lr){5-7} \cmidrule(lr){8-10} \cmidrule(lr){11-13}
Method & Pretraining & Train & Test & PSNR/SSIM$\uparrow$ & bpp$\downarrow$ & FPS$\uparrow$ & PSNR/SSIM$\uparrow$ & bpp$\downarrow$ & FPS$\uparrow$ & PSNR/SSIM$\uparrow$ & bpp$\downarrow$ & FPS$\uparrow$ \\
\midrule

NeRV & Per Video & $480$p & $480$p &
19.92 / 0.5511 & 0.0801 & 33.9 / 447.0 & 16.46 / 0.4424 & 0.0818 & 30.0 / \textbf{541.0} & 15.52 / 0.4887 & 0.1105 & 34.5 / \textbf{515.0} \\
HiNeRV & Per Video & $480$p & $480$p &
21.23 / 0.5915 & 0.0927 & 17.4 / 25.4 & 17.13 / 0.4607 & 0.0788 & 19.8 / 29.7 & 16.71 / 0.5065 & 0.1049 & 18.5 / 34.2 \\
NeRV-Enc* & Kinetics & $480$p & $480$p &
23.05 / 0.6491 & 0.1056 & 100.9 / \textbf{472.1} & 18.99 / 0.4694 & 0.1081 & 100.9 / 479.1 & 22.27 / 0.6234 & 0.1075 & 102.1 / 487.4 \\
TeCoNeRV\textsuperscript{1} (Ours) & Kinetics & $480$p & $480$p &
\textbf{25.52 / 0.7103} & \textbf{0.0676} & \textbf{167.0} / 465.2 & \textbf{20.65 / 0.4970} & \textbf{0.0705} & \textbf{176.0} / 450.5 & \textbf{24.21 / 0.6472} & \textbf{0.0729} & \textbf{166.3} / 472.6 \\

\midrule
NeRV & Per Video & $720$p & $720$p &
21.02 / 0.6402 & 0.0904 & 23.9 / \textbf{206.0} & 19.81 / 0.6684 & 0.0940 & 27.6 / \textbf{247.0} & 15.80 / 0.5187 & 0.1118 & 23.3 / \textbf{231.0} \\
HiNeRV & Per Video & $720$p & $720$p &
20.26 / 0.6437 & 0.0917 & 3.6 / 4.5 & 16.59 / 0.6340 & 0.0922 & 3.7 / 4.5 & 16.46 / 0.5427 & 0.1163 & 6.8 / 11.6 \\
NeRV-Enc* & Kinetics & $720$p &
$720$p & 20.28 / 0.6426 & 0.1030 & 20.1 / 187.4 & 18.22 / 0.6554 & 0.1027 & 20.3 / 194.1 & 19.46 / 0.5941 & 0.1029 & 20.32 / 195.9 \\

TeCoNeRV\textsuperscript{2} (Ours) & Kinetics & $720$p & $720$p &
25.22 / 0.7242 & 0.0667 & 60.9 / 155.8 &
24.33 / 0.7430 & 0.0580 & 63.2 / 150.0 &
24.10 / 0.6480 & 0.0736 & 61.0 / 144.8  \\
TeCoNeRV\textsuperscript{2} (Ours) & Kinetics & $480$p & $720$p &
\textbf{25.63 / 0.7380} & \textbf{0.0634} & \textbf{61.2} / 158.9 &
\textbf{24.95 / 0.7561} & \textbf{0.0514} & \textbf{67.3} / 173.4 &
\textbf{24.39 / 0.6578} & \textbf{0.0690} & \textbf{62.0} / 137.6 \\

\midrule

NeRV & Per Video & $1080$p & $1080$p &
21.32 / 0.6739 & 0.0827 & 13.6 / \textbf{101.0} & 18.52 / 0.5302 & 0.0858 & 16.5 / \textbf{119.9} & 16.47 / 0.5450 & 0.1209 & 15.3 / \textbf{117.4} \\
HiNeRV & Per Video & $1080$p & $1080$p &
20.30 / 0.6710 & 0.0656 & 2.0 / 2.6 & 17.47 / 0.5267 & 0.0994 & 2.1 / 2.9 & 14.73 / 0.5310 & 0.1366 & 2.3 / 3.8 \\
NeRV-Enc* & Kinetics & -- & $1080$p & -- & -- & -- & -- & -- & -- & -- & -- & -- \\
TeCoNeRV\textsuperscript{3} (Ours) & Kinetics & $720$p & $1080$p &
25.28 / 0.7274 & \textbf{0.0632} & 27.6 / 77.4 &
22.09 / 0.5682 & 0.0623 & 27.0 / 76.7 &
23.98 / 0.6572 & \textbf{0.0675} & 26.4 / 77.8 \\
TeCoNeRV\textsuperscript{3} (Ours) & Kinetics & $480$p & $1080$p &
\textbf{25.95 / 0.7389} & 0.0639 & \textbf{32.8} / 80.7 &
\textbf{22.62 / 0.5794} & \textbf{0.0622} & \textbf{33.1} / 84.6 &
\textbf{24.59 / 0.6644} & 0.0685 & \textbf{32.3} / 82.3 \\

\bottomrule
\end{tabular}
}
\end{table*}

\label{sub:impl_details}
\noindent\textbf{Implementation Details.} For NeRV-Enc*, we employ two variants of the hyponetwork, one for each of the $480$p and $720$p resolutions.
For NeRV and HiNeRV, we train a different network for each video at each resolution and report average metrics per test dataset.
For our method, we employ three configurations of non-overlapping patch tubelet models.
Specifically, we use patches of size $320{\times}160$ at $480$p, $320{\times}240$ at $720$p and $384{\times}270$ at $1080$p.
We denote these models as TeCoNeRV\textsuperscript{1, 2, 3} in \Cref{tab:allres_nonoverlap}. 
See \Cref{sec:supp_hypernet_details} for details on network settings.%

\subsection{Main Results}
\label{sub:main_res}

\begin{figure}[t]
\centering
\includegraphics[width=\linewidth]{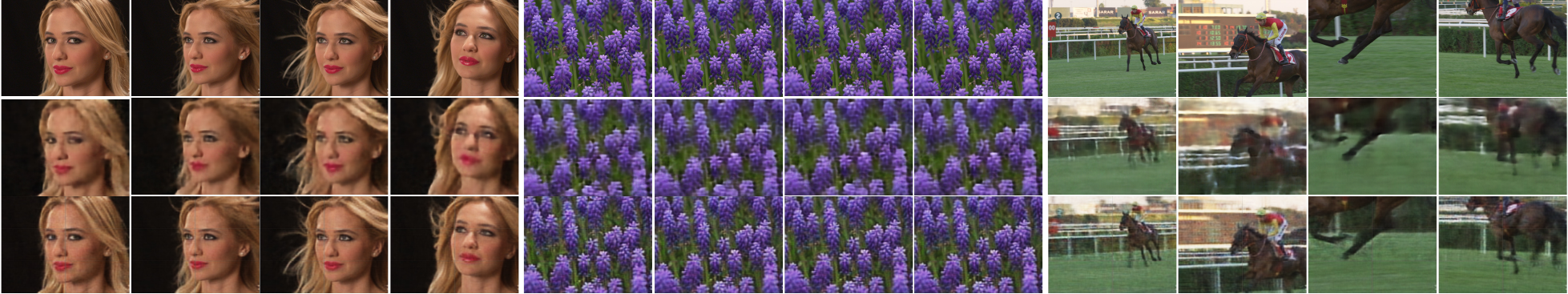}
\caption{
\textbf{Qualitative comparison at 480p on UVG.}
Left to right: Beauty, Honeybee and Jockey sequences from the UVG dataset.
Top to bottom: Ground truth, NeRV-Enc*, TeCoNeRV.
TeCoNeRV captures finer details like
facial features in Beauty, flowers in Honeybee and the horse’s legs in Jockey, which appear blurry in NeRV-Enc*.
Our method preserves structural integrity and important visual elements that are sometimes missing entirely in the NeRV-Enc* reconstruction.
}
\label{fig:qual_480p_nervenc_teco}
\end{figure}

\Cref{tab:allres_nonoverlap} presents video encoding performance on UVG, HEVC, and MCL-JCV across $480$p, $720$p and $1080$p resolutions.
At $480$p, TeCoNeRV achieves 25.52dB PSNR on UVG at 0.0676 bpp, a 2.47dB improvement over NeRV-Enc* (23.05dB at 0.1056 bpp) while encoding 1.65$\times$ faster, with consistent gains on HEVC and MCL-JCV.
At $720$p, the improvements are similarly substantial: 25.22dB versus 20.28dB on UVG, with our method encoding 3$\times$ faster.
Despite their fair encoding time allocations and bitrate budgets (\Cref{sub:baselines}), we also outperform NeRV and HiNeRV in terms of encoding speed, quality, and bitrate. \Cref{fig:qual_480p_nervenc_teco} compares the reconstructions from TeCoNeRV and NeRV-Enc* at 480p on UVG.
While the baseline produces blurry reconstructions with missing details, our model preserves structures and object integrity. %
At 720p (\Cref{fig:qual_720p_3_vid}), the degradation in the quality of NeRV-Enc* becomes more pronounced, consistent with the scaling limitation observed in \Cref{tab:allres_nonoverlap}.

For fair comparison, we train both NeRV-Enc* and our method on the same dataset splits -- using either the $480$p or $720$p Kinetics video subsets.
However, we also provide results with our models trained on $480$p videos for inference at $720$p and $1080$p resolutions.
This highlights a practical advantage of our patch-tubelet approach: models trained at lower resolutions can be applied to higher resolution inference, which we discuss further in \Cref{sub:res_indep}.
We observe a small boost in quality when using the 480p-trained models, which may be attributed to hyperparameters tuned at 480p and applied without re-tuning, and differences in video subsets and relative patch size. 
Encoding and decoding speeds are measured on an NVIDIA RTX A5000 GPU for $480$p and $720$p resolutions, and on RTX A6000 for $1080$p.

\begin{table}
\caption{\textbf{Comparison of number of parameters, wall time, and memory required for training NeRV-Enc* versus TeCoNeRV.}
Profiled on a single NVIDIA H200 GPU with batch size 1.}
\centering
\label{tab:train_time_mem}
\resizebox{0.7\linewidth}{!}{
\begin{tabular}{@{}l cc cc c c@{}}
\toprule
& & & \multicolumn{2}{c}{\# Params} & & \\
\cmidrule{4-5}
Method & Train & Test & Base & Unique & \makecell{Wall Time $\dagger$\\(hours)} & \makecell{Memory $\ddagger$\\(GB)} \\ 
\midrule
NeRV-Enc* & $480$p & $480$p & 436.8K & 96.3K & 28 & 6.5 \\
Ours & $480$p & $480$p & 65.2K & 15.9K & 6.0 + 3.5 & 1.8, 2.5 \\
Ours & $480$p & $720$p & 131.0K & 25.0K & 6.8 + 4 & 2.1, 2.9 \\
\midrule
NeRV-Enc* & $720$p & $720$p & 1.3M & 289.0K & 118 & 32 \\
Ours & $720$p & $720$p & 131.0K & 25.0K & 32 + 10 & 2.1, 2.9 \\
\bottomrule
\end{tabular}
}
\begin{minipage}
{0.68\linewidth}
\vspace{0.5mm}
\footnotesize
$\dagger$ Total wall time: 200 epochs (NeRV-Enc*); 150+50 epochs for ours (training + finetuning).\\
$\ddagger$ Memory: for training, then finetuning (ours only)
\end{minipage}
\end{table}

\noindent\paragraph{Memory and Scaling.}
NeRV-Enc's computational requirements reveal a fundamental scaling limitation (\Cref{tab:train_time_mem}).
Memory grows quadratically with frame size (or superlinearly with pixels), as the hypernetwork must predict a larger number of unique parameters, and also uses larger base parameters. 
With NeRV-Enc requiring 6.5GB at 480p and 32GB at 720p, training at 1080p is projected to require over 65GB (over 20$\times$ compared to our 2.9GB 1080p-capable model) at batch size 1, which makes training at this resolution slow and infeasible given our compute resources.
Furthermore, the $480$p$\to$$720$p scaling already demonstrates poor generalization in quality, with NeRV-Enc* achieving only 20.28dB on UVG at $720$p.
In contrast, our patch-tubelet decomposition maintains near-constant memory requirements regardless of resolution, enabling us to be the first hypernetwork-based approach to demonstrate results at resolutions $720$p and $1080$p, while previous methods report only at 256${\times}$256.
While we profile on a larger H200 GPU to accommodate NeRV-Enc, our models train efficiently with batch size 32 on more accessible hardware like 4 RTX A4000 GPUs.

\begin{table}
\centering
\caption{
\textbf{Comparison of TeCoNeRV and NeRV-Enc* at 480p with ablations on residual encoding and patch-tubelets.}
Results reported on UVG and Kinetics-400 at $480$p resolution. 
Our methods achieve superior quality while using significantly fewer parameters. 
Residual encoding (Res.) improves bpp and encoding/decoding speeds (FPS) across all models. 
}
\label{tab:detail_nervenc}
\renewcommand{\tabcolsep}{3pt}
\resizebox{0.9\linewidth}{!}{%
\begin{tabular}{@{}l cc ccc ccc@{}}
\toprule
  &  &  & \multicolumn{3}{c}{UVG~\cite{mercat2020uvg}} & \multicolumn{3}{c}{Kinetics-400~\cite{kay2017kinetics}} \\
\cmidrule(l){4-6}
\cmidrule(l){7-9}
Method & Res. & $\mathcal{L}_{\text{temp}}$
& PSNR / SSIM $\uparrow$ & bpp $\downarrow$ & FPS $\uparrow$ 
& PSNR / SSIM $\uparrow$ & bpp $\downarrow$ & FPS $\uparrow$ \\
\midrule
NeRV-Enc* & \xmark & -- & 23.05 / 0.6491 & 0.1056 & 101 / 472 & 22.86 / 0.7208 & 0.1078 & 99 / 459 \\
NeRV-Enc* & \cmark & -- & 23.48 / 0.6608 & 0.0649 & 284 / 544 & 23.17 / 0.7305 & 0.0698 & 275 / 543 \\
Ours & \xmark & \xmark & 24.92 / 0.6985 & 0.1045 & 89 / 315 & 24.84 / 0.7495 & 0.1073 & 89 / 313 \\
Ours & \cmark & \xmark & 25.49 / 0.7159 & 0.0826 & 125 / 337 & 25.28 / 0.7638 & 0.0861 & 125 / 329 \\
TeCoNeRV & \cmark & \cmark &
25.52 / 0.7103 & 0.0676 & 167 / 465 &
25.15 / 0.7208  & 0.0678 & 171 / 466 \\
\bottomrule
\end{tabular}
}
\end{table}

\begin{figure}[htb]
\centering
\includegraphics[width=0.85\linewidth]{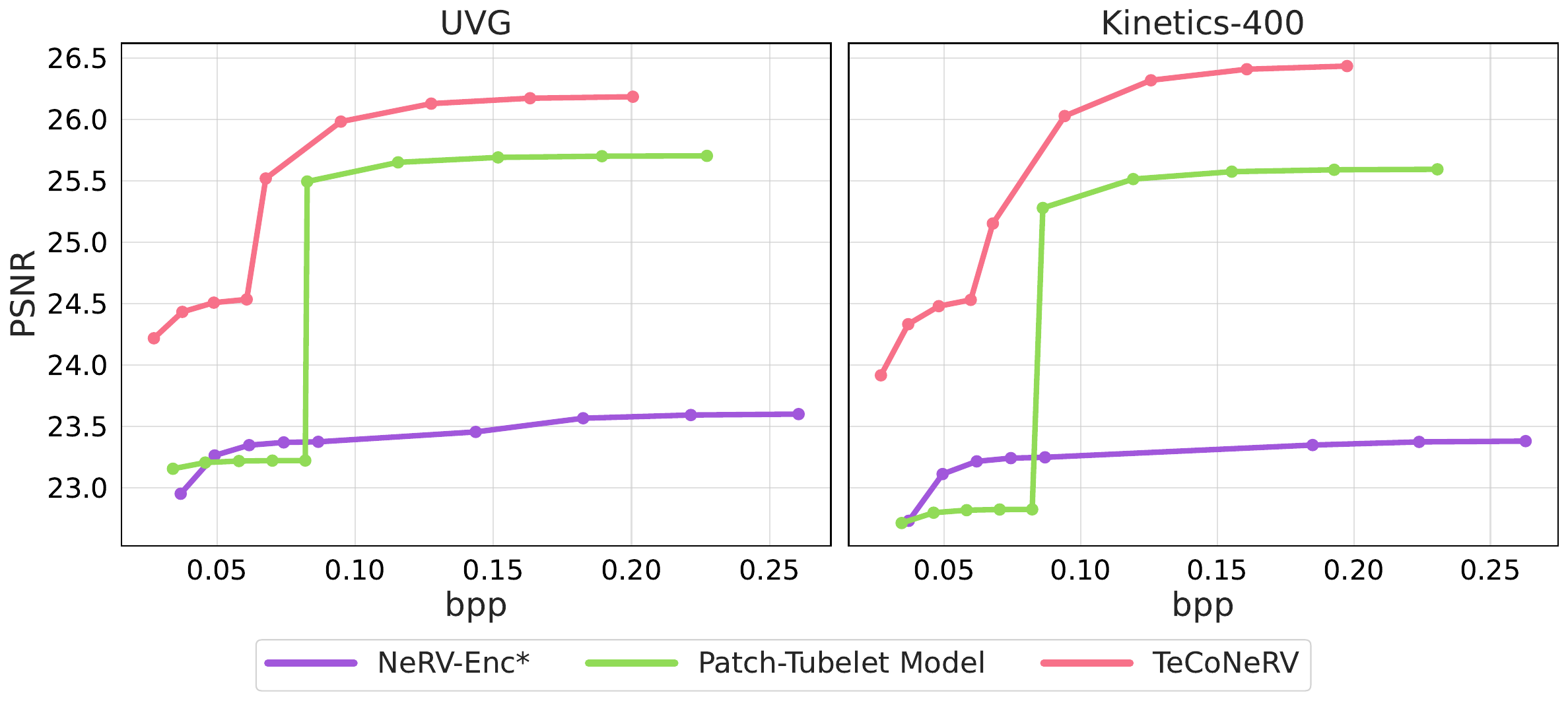}
\caption{
\textbf{Quality vs. bitrate} on UVG (left) and Kinetics-400 (right) at 480p. 
Rate-distortion curves showing PSNR vs. bpp for NeRV-Enc*, Patch-Tubelet and TeCoNeRV models. 
For each method, we show the Pareto frontier across two model size configurations.
TeCoNeRV outperforms the baseline, achieving up to 2.47dB higher PSNR at comparable bitrates.
}
\label{fig:rd_plot_nervenc}
\end{figure}

\noindent\textbf{Impact of residual encoding.} \Cref{tab:detail_nervenc} examines our design choices in detail at $480$p, evaluated on both UVG and Kinetics-400.
We compare NeRV-Enc* against our patch-tubelet approach, with and without temporal coherence regularization.
Residual encoding consistently improves compression efficiency across all methods while accelerating both encoding and decoding.
Our Patch-Tubelet approach demonstrates improvement over the baseline even without temporal coherence regularization, achieving 25.49dB PSNR compared to NeRV-Enc*'s 23.48dB when both use residual encoding.
Adding temporal coherence regularization yields TeCoNeRV, which maintains comparable reconstruction quality while reducing bitrate by an additional 18\%, and achieving the highest encoding speed.
\Cref{fig:rd_plot_nervenc} illustrates the rate-distortion Pareto frontier for each method, using two size configurations across varying quantization levels (4-8 bits).
TeCoNeRV consistently outperforms the baseline throughout the operating range.

\subsection{Visual Quality versus Size}
\label{sub:quality_vs_size}

\begin{figure}[t]
\centering
\includegraphics[width=\linewidth]{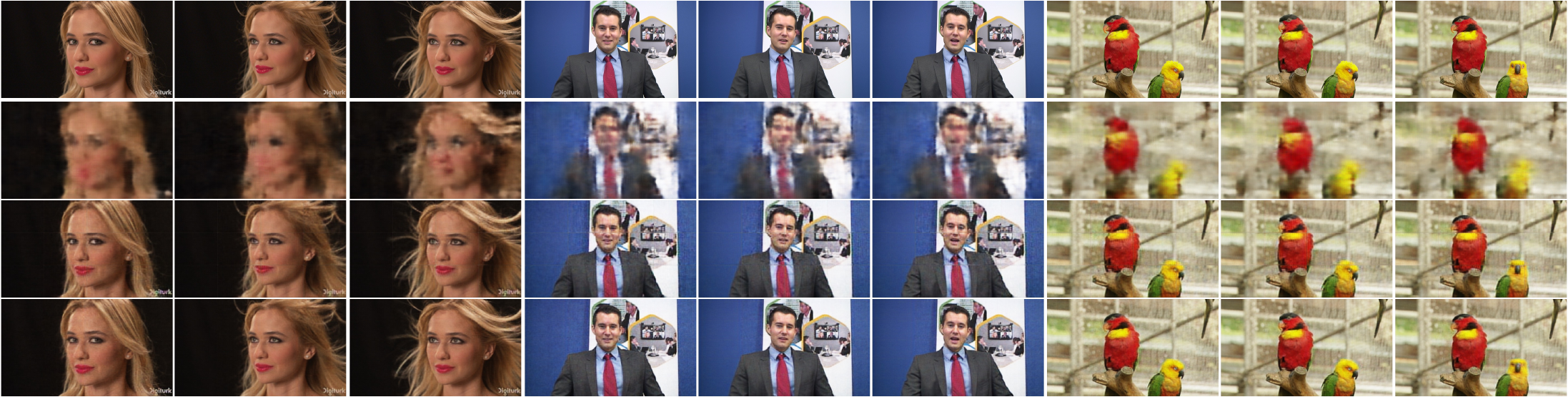}
\caption{\textbf{Qualitative comparison at 720p.}
We show 3 frames from UVG, HEVC Class E, and MCL-JCV datasets. 
Top to bottom: Ground truth, NeRV-Enc*, ours (``no overlap''), ours (``overlap with cropping''). 
Our method's reconstructions preserve structural details, while the baseline exhibits noticeable quality degradation.
Our ``overlap with cropping'' patch fusion strategy eliminates the boundary artifacts in ours (``no overlap''). 
Best viewed digitally and zoomed in.
}
\label{fig:qual_720p_3_vid}
\end{figure}

Our patch-based design may produce boundary artifacts, which can become more pronounced at higher resolutions.
We propose to mitigate these artifacts by using strided patch-tubelets with overlap at inference.
As our primary fusion strategy, we propose cropping to remove redundant edges from adjacent patches before tiling.
In this way, we can extend a model trained for $480$p inference to infer at $720$p or $1080$p, even when the patch size does not exactly divide the target resolution.
\Cref{fig:qual_720p_3_vid} demonstrates that this approach effectively eliminates artifacts from non-overlapping patches.
Additional qualitative results at $1080$p are included in \Cref{fig:supp_qual_1080p}. %

\begin{table}
\centering
\caption{
\textbf{Rate control via patch configuration on UVG}.
Representative configurations with and without overlap, demonstrating quality-bitrate trade-offs at 480p, 720p, and 1080p inference.
}
\label{tab:overlap_patches}
\renewcommand{\tabcolsep}{4pt}
\resizebox{0.65\linewidth}{!}{%
\begin{tabular}{@{}cc cc cc@{}}
\toprule
Train & Test &
Overlap & Patch Size & PSNR / SSIM $\uparrow$ & bpp $\downarrow$ \\
\midrule

$480$p & $480$p & \xmark & $320{\times}160$ & \textbf{25.52 / 0.7103} & \textbf{0.0676} \\
$480$p & $480$p & $\cmark$ & $256{\times}192$ & 25.28 / 0.7022 & 0.1048 \\
\midrule
$720$p & $720$p & \xmark & $320{\times}240$ & 25.22 / 0.7242  & \textbf{0.0667} \\
$480$p & $720$p & $\cmark$ & $320{\times}160$ & \textbf{26.45 / 0.7533} & 0.0985 \\

\midrule
$720$p & $1080$p & \xmark & $384{\times}270$ & 25.28 / 0.7274	& \textbf{0.0632} \\
$480$p & $1080$p & $\cmark$ & $320{\times}240$ & \textbf{26.89 / 0.7572} & 0.0850 \\

\bottomrule
\end{tabular}%
}
\end{table}

\begin{figure}[htb]
\centering
\includegraphics[width=0.75\linewidth]{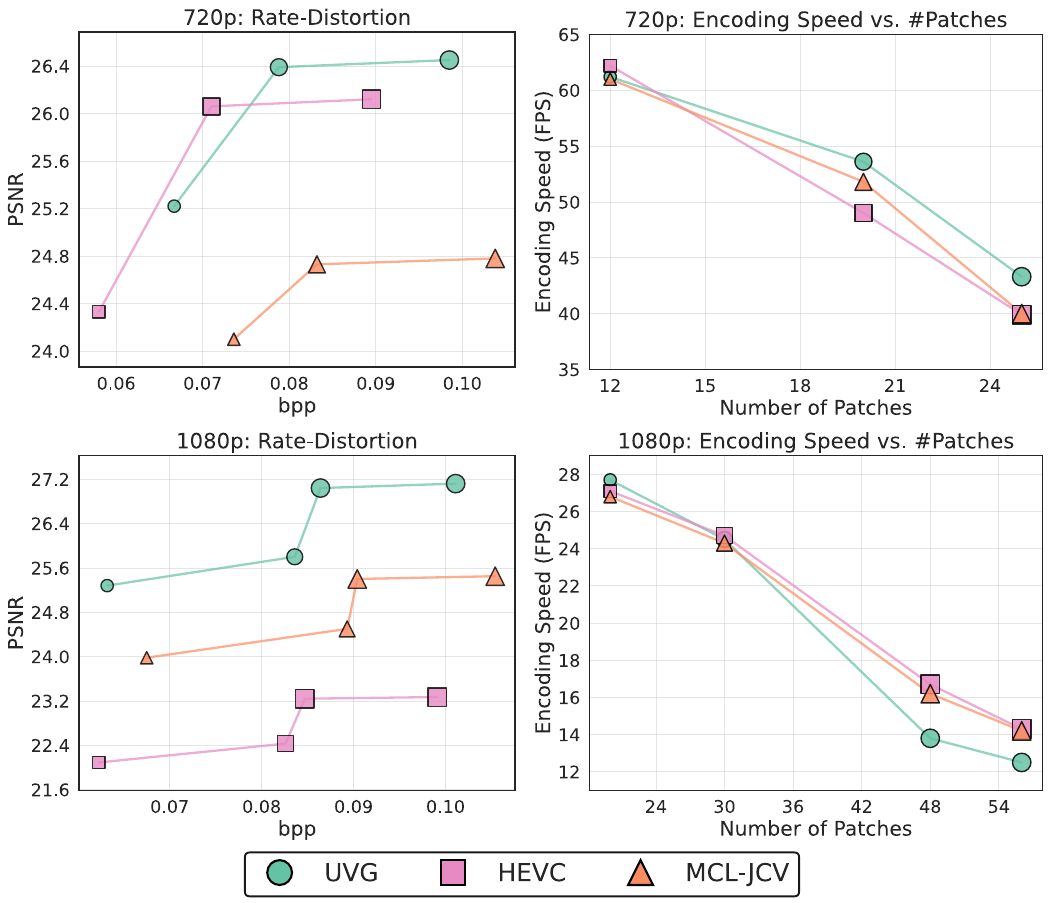}
\caption{\textbf{Rate-distortion and encoding speed vs. number of patches.}
Varying patch size and overlap produces different operating points at 720p (top) and 1080p (bottom) inference.
Marker size corresponds to number of patches.
}
\label{fig:overlap_patches_plots}
\end{figure}

\begin{figure}[bt]
\centering
\includegraphics[width=0.95\linewidth]{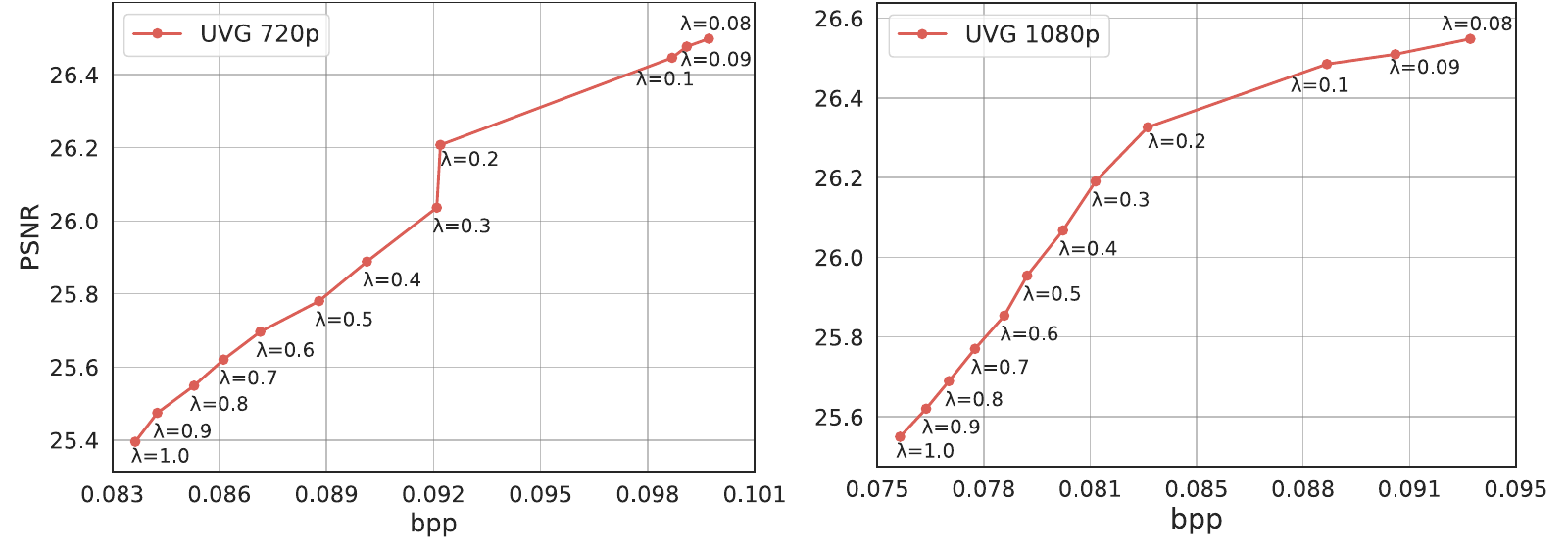}
\caption{\textbf{Rate control with temporal regularization.} Finetuning our method with varying amounts of regularization ($\lambda_\text{temp}$) allows us to reliably trade off PSNR and bpp, offering interpretable rate control. Left: $320{\times}160$ model (trained at $480$p, inferred at $720$p with overlap) on UVG, right: $320{\times}240$ model (trained at $720$p, inferred at $1080$p with overlap).
}
\label{fig:l1_lever_overlap_combined}
\vspace{-1em}
\end{figure}

\definecolor{480p_infer}{HTML}{fc8d62}
\definecolor{720p_infer}{HTML}{e78ac3}
\definecolor{1080p_infer}{HTML}{66c2a5}
\begin{figure}[t]
\begin{minipage}{\linewidth}
\centering
\captionsetup{type=table}
\caption{
\textbf{Resolution-independent training on UVG.}
A single model trained at 480p achieves competitive quality at 480p, 720p, and 1080p.
Stars ($\bigstar$) and squares ($\blacksquare$) denote non-overlapping patches, other markers show overlap configurations.
}
\label{tab:generalize_overlap_patches}
\resizebox{0.78\linewidth}{!}{
\begin{tabular}{@{}c c c cc ccc@{}}
\toprule
& Train & Test & Overlap & Patch Size & PSNR / SSIM $\uparrow$ & bpp $\downarrow$ \\
\midrule
\textcolor{480p_infer}{$\bigstar$} & $480$p & $480$p & \xmark & $320{\times}160$ &
25.52 / 0.7103 & 0.0676 \\
\textcolor{720p_infer}{$\bigstar$} & $480$p & $720$p & \xmark & $320{\times}240$ &
25.63 / 0.7380 & 0.0634 \\
\textcolor{1080p_infer}{$\bigstar$} & $480$p & $1080$p & \xmark & $384{\times}270$ &
25.95 / 0.7389 & 0.0639 \\

\midrule

\textcolor{720p_infer}{\scalebox{1.8}{$\bullet$}} & $480$p & $720$p & $\cmark$ & $320{\times}160$ &
26.45 / 0.7533 & 0.0985 \\
\textcolor{1080p_infer}{\scalebox{1.8}{$\bullet$}} & $480$p & $1080$p & $\cmark$ & $320{\times}160$ &
27.12 / 0.7601 & 0.1011 \\
\textcolor{1080p_infer}{\scalebox{1.8}{$\bullet$}} & $480$p & $1080$p & $\cmark$ & $320{\times}240$ &
26.89 / 0.7572 & 0.0850 \\

\midrule
\textcolor{720p_infer}{$\blacksquare$} & $720$p & $720$p & \xmark & $320{\times}240$ &
25.22 / 0.7242 & 0.0667 \\
\textcolor{1080p_infer}{$\blacksquare$} & $720$p & $1080$p & \xmark & $384{\times}270$ &
25.28 / 0.7274 & 0.0632 \\

\midrule
\textcolor{1080p_infer}{\scalebox{1.2}{$\blacktriangle$}} & $720$p & $1080$p & $\cmark$ & $320{\times}240$ &
26.48 / 0.7479 & 0.0887 \\

\bottomrule
\end{tabular}
}
\end{minipage}

\vspace{1em}

\begin{minipage}{\linewidth}
\centering
\includegraphics[width=0.8\linewidth]{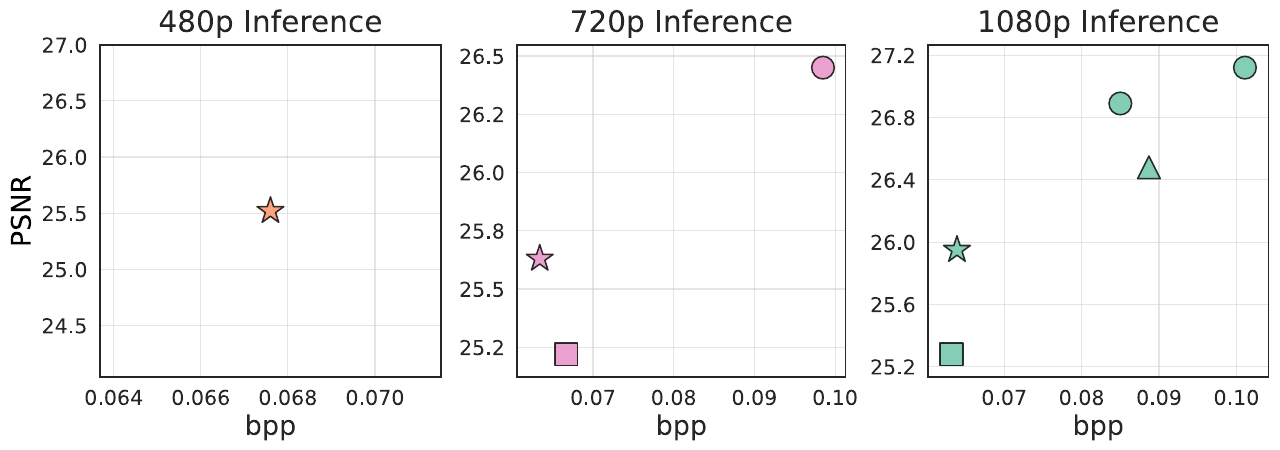}
\captionsetup{type=figure}
\caption{
\textbf{Our models trained at lower resolutions scale for prediction at higher resolutions.}
Rate-distortion curves showing competitive performance of models trained on lower-resolution videos at all target resolutions (settings in Table~\ref{tab:generalize_overlap_patches}).
}
\label{fig:generalize_overlap_plot}
\end{minipage}
\end{figure}

However, more patches may increase bitstream size, necessitating rate control.  
Our method provides two complementary mechanisms.
First, to control patch count, we adjust patch size and overlap when scaling resolutions.
For instance, while our $320{\times}160$ model trained at $480$p can scale to $1080$p, using the larger $320{\times}240$ model trained at $720$p is more practical, producing fewer patches and a better quality-size balance.
\Cref{tab:overlap_patches} shows representative operating points with and without overlap at each resolution.
\Cref{fig:overlap_patches_plots} illustrates the full trade-off space.
Second, temporal coherence strength $\lambda_\text{temp}$ provides direct rate control during finetuning.
\Cref{fig:l1_lever_overlap_combined} demonstrates that 
by varying $\lambda_\text{temp}$, we can achieve up to 18\% reduction in bpp at the cost of only 4\% PSNR loss, offering an interpretable mechanism for balancing quality with bitrate.

\subsection{Resolution-Independent Training}
\label{sub:res_indep}
Combining our patch-tubelet design with the overlap strategies in \Cref{sub:quality_vs_size} enables models trained at one resolution to infer at higher resolutions.
\Cref{tab:allres_nonoverlap} illustrates this capability: models trained at $480$p achieve competitive quality at $720$p and $1080$p
without overlap. Similarly, models trained on $720$p scale to $1080$p.
This allows us to train models to encode at $1080$p even in the absence of high-resolution uncompressed pre-training data, which is scarce.
Additionally, training a model for $720$p or $1080$p inference on $480$p videos significantly speeds up training time.
\Cref{tab:generalize_overlap_patches} and \Cref{fig:generalize_overlap_plot} further illustrate this flexibility: a single $480$p-trained model achieves strong performance across all resolutions by varying patch size and overlap at inference.
Similarly, a model trained on $720$p videos can infer at $1080$p, and this becomes applicable to even higher resolutions, highlighting our unique paradigm.
The hypernetwork learns to predict weights for fixed-size spatio-temporal regions, making resolution scaling a function of patch count rather than model capacity.
This is fundamentally impossible for NeRV-Enc, which can only predict videos of a fixed resolution and whose memory constraints prevent high-resolution training. 

\subsection{Encoding Strategies}
\label{sub:ablations}

\begin{table}
\centering
\caption{\textbf{Impact of residual encoding at different quantization levels}. We compare three encoding approaches: storing full parameters for each clip (No Residuals), residuals relative to the first clip, and residuals relative to the previous clip. Results use our TeCoNeRV model with temporal coherence regularization. At higher compression rates, residual encoding significantly outperforms direct parameter storage, with previous-clip residuals providing the best compression efficiency.}
\label{tab:ablate_residuals}
\renewcommand{\tabcolsep}{3pt}
\resizebox{0.9\linewidth}{!}{
\begin{tabular}{@{}c cc cc cc@{}}
\toprule
 & \multicolumn{2}{c}{No Residuals} & \multicolumn{2}{c}{Residual from First} & \multicolumn{2}{c}{Residual from Previous} \\
\cmidrule(l){2-3}
\cmidrule(l){4-5}
\cmidrule(l){6-7}
\makecell{Quant Level \\ (bits)}  & PSNR/SSIM $\uparrow$ & bpp $\downarrow$ & PSNR/SSIM $\uparrow$ & bpp $\downarrow$ & PSNR/SSIM $\uparrow$ & bpp $\downarrow$ \\
\midrule
8 & 26.04 / 0.7281 & 0.2324 & 26.18 / 0.7318 & 0.2184  & 26.18 / 0.7318 & 0.2005 \\
7 & 25.64 / 0.7174 & 0.1946 & 26.17 / 0.7313 & 0.1863 & 26.17 / 0.7314 & 0.1633 \\
6 & 24.39 / 0.6796 & 0.1567 & 26.10 / 0.7295 & 0.1578 & 26.13 / 0.7301 & 0.1275 \\
5 & 21.53 / 0.5774 & 0.1202 & 25.89 / 0.7228 & 0.1347  & 25.98 / 0.7253 & 0.0948 \\
4 & 17.26 / 0.4061 & 0.0879 & 25.03 / 0.6962 & 0.1136  & 25.52 / 0.7103 & 0.0676 \\
\bottomrule
\end{tabular}
}
\end{table}

\Cref{tab:ablate_residuals} compares compression efficiency across three encoding strategies: storing complete parameters for each clip, residuals relative to the first clip and residuals relative to the previous clip, across quantization levels. 
At aggressive quantization, direct parameter storage degrades significantly in quality, while residual encoding maintains robust performance.
Thus, residual encoding not only improves compression efficiency but also enhances resilience to quantization, with previous-clip residuals providing the best quality-compression trade-off.
Further, we note that our previous-clip residuals can also handle scene changes and dynamic motion robustly. See \Cref{sec:supp_concat_uvg} for further details.
For applications requiring random access, first-clip residuals provide direct access to any clip, or alternatively, ``keyframe" clips with full parameters can be inserted periodically in the previous-clip strategy.
Additional ablations on patch sizes and hyponetwork parameter distributions are provided in \Cref{sec:supp_ablation}.

\section{Conclusion}
\label{sec:conclusion}
We address fundamental limitations of existing INR-based video compression methods. 
By decomposing videos into patch tubelets, we enable hypernetwork-based approaches to scale to higher resolution videos, overcoming the memory constraints that limited prior methods. 
We combine differential encoding with temporal coherence regularization to efficiently store only small residual weights between consecutive clips.
Together, these contributions change the scaling behavior of hypernetwork INR compression: from sparsely sampled frames at low resolution to full high-resolution sequences, at near-constant training memory and smaller compressed size.
Further, our framework enables resolution-independent training, allowing models trained at lower resolutions to operate at higher resolutions. 
Notably, our method maintains fast encoding speeds of previous hypernetwork-based approaches, demonstrating that hypernetwork methods can achieve both scalability and efficiency for practical video compression.
We hope our work motivates further research on improving reconstruction quality and compressed size within the hypernetwork representation class.

\newpage

\noindent{\textbf{Acknowledgements.}} We would like to thank Pulkit Kumar for his valuable guidance on scaling our experimental pipeline, and Max Ehrlich for the insightful technical discussions. This work was partially supported by the NSF CAREER Award (\#2238769) to AS. This research is based upon work supported by the Office of the Director of National Intelligence (ODNI), Intelligence Advanced Research Projects Activity (IARPA), via IARPA R\&D Contract No. 140D0423C0076. The views and conclusions contained herein are those of the authors and should not be interpreted as necessarily representing the official policies or endorsements, either expressed or implied, of the NSF, ODNI, IARPA, or the U.S. Government. The U.S. Government is authorized to reproduce and distribute reprints for Governmental purposes notwithstanding any copyright annotation thereon.

\FloatBarrier

\FloatBarrier

\bibliography{main}

\bmvasupplementarytitle
\addtocontents{toc}{\protect\setcounter{tocdepth}{2}}

\setcounter{page}{1}

\suppltableofcontents
\appendix
\crefalias{section}{appendix}
\crefalias{subsection}{appendix}

\section{Future Work}
\label{sec:future_work}
While our encoding speeds substantially improve over prior hypernetwork methods (\Cref{tab:allres_nonoverlap}), decoding speeds at higher resolutions (1080p) with increased number of patches can be further optimized, to match that of NeRV.
Parallel decoding (forward pass through hyponetwork) of patches across multiple GPUs, along with other simple optimizations, offers a promising direction to further reduce decoding latency.

Further, visual quality and bitrate still fall below established traditional and autoencoder-based codecs.
However, TeCoNeRV represents an important advancement in efficiently scaling implicit neural video compression to higher resolution content using hypernetworks. 
Future work could explore improved hyponetwork architectures with grid-based positional encoding.
Additionally, our results are reported on an initial setup using 10,000 training videos from Kinetics.
In practice, training only once on a large, diverse video corpus would yield significant further gains.
Acquiring large-scale high-resolution training video data and scaling the hypernetwork transformer remains an open challenge.

\section{Additional Dataset Details}
We primarily evaluate on the UVG~\cite{mercat2020uvg}, HEVC~\cite{ohm2012comparison,sullivan2012overview} and MCL-JCV~\cite{wang2016mcl} datasets.
The UVG dataset consists of 7 sequences, each of 600 frames in length except for ShakeNDry, which has 300 frames. All sequences are of resolution $1920{\times}1080$.
MCL-JCV comprises 24 sequences between 120-150 frames in length, with varying amounts of motion. These sequences are provided at both $1280{\times}720$ and $1920{\times}1080$.
The HEVC standard test sequences follow the Common Test Conditions (CTC) established by ITU-T and ISO/IEC JCT-VC for benchmarking video compression algorithms.
This dataset consists of five classes, A-E. We omit evaluation on classes A and D since they consist of videos significantly larger than 1080p ($2560{\times}1600$) or smaller than 480p ($416{\times}240$). 
We report on Classes B, C and E, which comprise 5 videos of size $1920{\times}1080$, 4 videos of size $832{\times}480$ and 3 videos of size $1280{\times}720$ respectively. These videos vary between 240-600 frames in length and examples include Traffic, PeopleOnStreet, Kimono and BasketballDrive.
NeRV-Enc~\cite{chen2024fast} further reports on Kinetics-400. Hence, we compare against this baseline on a subset of Kinetics validation videos at 480p, consisting of one video per class.
Videos are preprocessed by downsampling to the target inference resolution by resizing to 480, 720 or 1080 pixels on the shorter side with anti-aliasing and preserved aspect ratio, followed by center cropping to the exact dimensions.

\section{Compression Pipeline}

Our compression pipeline consists of two stages: uniform quantization followed by lossless entropy coding. 

\noindent\textbf{Quantization.} 
Following prior work on video INR compression~\cite{chen2021nerv}, we apply uniform quantization to reduce precision of model weights, significantly reducing the final model size. 
We quantize both the initial clip hyponetwork weights and subsequent residuals to varying bit depths (between 4-8 bits).
While 8-bit quantization introduces negligible quality degradation, more aggressive quantization levels enable further compression at the cost of reconstruction quality, as demonstrated in \Cref{tab:ablate_residuals}. %
We report bpp at 4 bits by default in our work, unless otherwise specified.

\noindent\textbf{Entropy Coding.} 
After quantization, we apply arithmetic coding using the torchac library~\cite{mentzer2019practical} to losslessly compress the quantized weights. 
This produces the final bitstream, from which we compute all reported bits-per-pixel (bpp) values. Quality metrics (PSNR/SSIM) are measured after loading the compressed model and decompressing it. 
In \Cref{supp_subsec:huffman}, we provide an ablation comparing arithmetic coding to Huffman coding, finding that arithmetic coding achieves marginally better compression ratios.

\noindent\textbf{Rate-Distortion Curves.} 
To generate the rate-distortion curves in \Cref{fig:rd_plot_nervenc}, we train two model configurations per method -- one with larger hyponetwork size and one with smaller hyponetwork size (detailed in \Cref{sec:supp_hypernet_details}). 
We quantize each configuration at 4, 5, 6, 7, and 8 bits and compress with arithmetic coding. 
We plot quality-bitrate pairs in strictly ascending order of bitrate, forming a Pareto front: that is, if a higher-bitrate configuration achieves lower quality than a lower-bitrate one, we exclude it from the curve. 
This can occur, for example, when transitioning from an 8-bit quantized small model to a 4-bit quantized large model.

\section{Robustness to Scene Cuts and Dynamic Content}
\label{sec:supp_concat_uvg}
\vspace{-0.5em}
Our default residual encoding scheme stores full weights only for the first clip and previous-clip residuals thereafter.
Here, we evaluate the robustness of previous-clip residuals to abrupt changes in content as well as dynamic scenes. 
We concatenate all 7 sequences from UVG 1080p into a single video of 3900 frames in the following order: Beauty, ReadySteadyGo, HoneyBee, Jockey, YachtRide, Bosphore, and ShakeNDry.
This introduces 6 hard shot changes alongside diverse content within and across sequences, and includes sequences with fast motion, such as Jockey and ReadySteadyGo. %
We encode this concatenated video using TeCoNeRV with previous-clip residuals, and without inserting any key-clips.

\begin{figure}[ht]
\centering
\includegraphics[width=\linewidth]{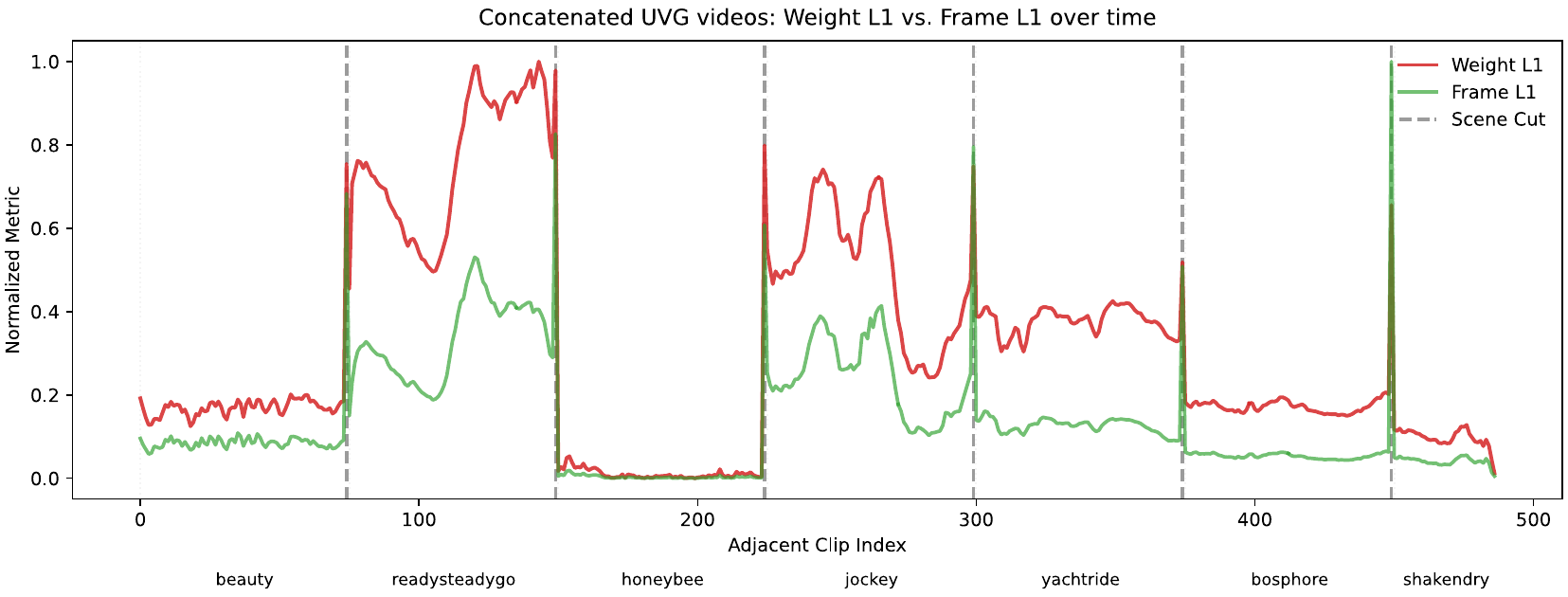}
\caption{\textbf{Weight L1 and frame L1 across the concatenated UVG sequence.} We encode all 7 UVG sequences concatenated into a single 3900-frame 1080p video, using previous-clip residuals. Normalized weight-space L1 (red) and frame-space L1 (green) are plotted over adjacent clip pairs; dashed lines mark scene boundaries between sequences. Weight L1 spikes at each scene cut and in high-motion regions within sequences, closely tracking the frame-space signal. This indicates that the residual scheme responds to content change.}
\label{fig:concat_uvg}
\end{figure}

Our method achieves a reconstruction PSNR of 26.07 dB with 0.7427 SSIM at 0.0629 bpp, which is consistent with per-video encoding (\Cref{tab:allres_nonoverlap}), without degradation.
This consistency holds because our residual encoding has a built-in self-correction property: each clip's residual is computed against the reconstructed (dequantized) state of the previous clip, and not the raw prediction.
Thus, even when the residual between consecutive clips may become large at scene boundaries and during fast motion, these large transitions are captured locally and encoded robustly, minimizing error propagation to subsequent clips.
\Cref{fig:concat_uvg} shows that the spikes in the $\ell_1$-norm of the weight-space align with scene boundaries and high-motion regions within sequences, mirroring the frame-space $\ell_1$ signal, computed as the mean absolute pixel difference between consecutive frames.
The previous-clip residual scheme therefore robustly adapts to content changes, allocating a higher bitrate where content genuinely changes and remaining compact where it does not.
We report on standard video compression benchmarks and leave the investigation of adaptive key-clip insertion for longer videos to future work. This also supports the design of $\mathcal{L}_{\text{temp}}$ (\Cref{sec:temp_coh}). Although \Cref{eqn:temp_loss} is not explicitly content-weighted, the weights it penalizes are themselves predicted from the input clips, through $\theta=g_\phi(S)$. Thus, weight-space distance tracks pixel-space distance without being conditioned on it; and temporal proximity proves to be a good prior for pixel similarity, on average.

\section{Additional Details on NeRV and HiNeRV}
\label{sec:supp_nerv_hinerv_details}
Frame-wise video INRs compress videos by overfitting a neural network on each video individually, predicting frames from spatial or temporal coordinates. 
The original NeRV~\cite{chen2021nerv} uses convolution-based upsampling blocks with sinusoidal positional encoding, to predict a frame at timestep $t$.
HiNeRV~\cite{kwan2023hinerv} introduces several improvements including grid-based positional encodings, hierarchical skip connections, patchwise processing, and bilinear upsampling with ConvNext blocks, achieving state-of-the-art reconstruction quality among frame-wise INR methods. Note that we use the improved NeRV architecture from HNeRV~\cite{chen2023hnerv} and follow their scaling rules for network configuration. 
For HiNeRV~\cite{kwan2023hinerv}, we adapt settings from the paper to our resolutions and bitrates, tuning base size, patch size, and $fc_\text{dim}$ to obtain best settings for target compression rates. 

\noindent\textbf{Training protocol.} 
These per-video optimization methods typically require training for hundreds of epochs per video, with state-of-the-art models requiring hours of encoding time.
Thus, for these models, encoding time includes the per-video training time.
To establish fair comparison, we configure both baselines with target bpp at least as high as our model and encoding time budgets longer than our hypernetwork inference time. We train NeRV for 5 epochs per video (fewer epochs produce insufficient quality for meaningful comparison) and HiNeRV for 1 epoch per video. Quality and bpp are computed per video and averaged per dataset.
While these methods can achieve better PSNR given several hours of per-video optimization, hypernetworks provide a generalizable alternative that alleviates the optimization bottleneck in scenarios that require fast encoding.
We note that the quality of NeRV and HiNeRV reported here do not reflect their peak operating point, but rather their operating point under budgets established for fair cross-regime comparison (\Cref{sub:baselines}).

\noindent\textbf{Network configurations.} 
We use the following settings for each 600-frame UVG sequence.
For NeRV, we vary \texttt{modelsize}, $fc_{\text{hw}}$, and decoder strides across resolutions:
\begin{itemize}
    \item 480p: \texttt{modelsize} = 2.4, strides = [5, 2, 2, 2], $fc_{\text{hw}} = (12, 16)$

    \item 720p: \texttt{modelsize} = 9.0, strides = [5, 2, 2, 2, 2],  $fc_{\text{hw}} = (9, 16)$

    \item 1080p: \texttt{modelsize} = 16.0, strides = [5, 3, 2, 2, 2], $fc_{\text{hw}} = (9, 16)$
\end{itemize}

For HiNeRV, we adjust $fc_{\text{dim}}$, the base grid configuration, and patch sizes:
\begin{itemize}
    \item 480p: $fc_{\text{dim}} = 230$, \texttt{base\_grid\_size} = [75, 12, 16, 2], patch size = $80{\times}80$

    \item 720p: $fc_{\text{dim}} = 512$, \texttt{base\_grid\_size} = [150, 18, 32, 2], patch size = $80{\times}80$

    \item 1080p: $fc_{\text{dim}} = 556$, \texttt{base\_grid\_size} = [150, 18, 32, 2], patch size = $120{\times}120$
        (This is the largest HiNeRV configuration we were able to fit on an RTX A6000 GPU at 1080p.)
\end{itemize}

Further, we scale \texttt{modelsize} (NeRV) and $fc_{\text{hw}}$ (HiNeRV) to accommodate varying video lengths in HEVC and MCL-JCV datasets.
While NeRV achieves slightly faster decoding speeds than our hypernetwork approach (\Cref{tab:allres_nonoverlap}), it does not match TeCoNeRV's reconstruction quality at comparable bitrates and requires substantially longer encoding times. 
We leave investigation of faster decoding within the hypernetwork paradigm to future work, as discussed in \Cref{sec:future_work}.

\section{Details of the Hypernetwork and Hyponetwork Architecture}
\label{sec:supp_hypernet_details}
We provide configuration details for all models evaluated in the main paper.
Our architecture is based on the hypernetwork-hyponetwork framework from \cite{chen2024fast}, which we briefly review before detailing configurations of the models we implement. 

\subsection{Framework Overview}
In this architecture, a hypernetwork predicts the weights of a hyponetwork (HypoNeRV) for each input video clip.
The HypoNeRV consists of multiple NeRV blocks~\cite{chen2021nerv}, each containing convolutional layers followed by PixelShuffle~\cite{shi2016real} upsampling.
This network takes a time positional encoding as input and upsamples it until it reaches the target spatial resolution.
Our implementations use 4-layer (ours) or 5-layer (NeRV-Enc*) architectures with learnable convolutional parameters at each layer.
Following~\cite{chen2024fast}, we maintain consistent convolutional channel width except for the input time position embedding (PE) and output layer (3 RGB channels).

The learnable parameters of this system are divided into base parameters and unique parameters.
The base parameters encompass weights and biases across the layers of the hyponetwork, capturing common patterns and representational capacity shared across diverse video content, while unique parameters encode clip-specific information.
This separation enables efficient adaptation to individual clips while maintaining a shared representational foundation.

To predict the unique parameters for a clip, the hypernetwork processes two types of tokens -- data tokens representing the tokenized input clip, and a set of learnable weight tokens corresponding to layers of the hyponetwork.
The Transformer backbone processes these jointly through self-attention and outputs the encoded weight tokens.
These are projected via fully connected (FC) layers to convert from the transformer token dimension to the weight token dimension.
The tokens are then repeated, and element-wise multiplied with base parameters, followed by normalization to yield the final modulated hyponetwork weights for the input clip (\Cref{fig:method}).

In practical deployment, the hypernetwork serves as the encoder and the base parameters are installed in the decoder.
Base parameters need not be transmitted per clip, they are learned once and shared across all videos.
Therefore, we exclude base parameters from bitrate (bpp) calculations, considering only unique parameters and residuals.

\subsection{Training Settings}
All models train for 150 epochs followed by finetuning for 50 epochs (with or without temporal coherence regularization).
We use the Adam optimizer with a learning rate of $1e-4$ across experiments.
Input videos are organized into clips of 8 consecutive frames.
This approach differs from \cite{chen2024fast}, which only encodes sparsely sampled frames from videos. 
During training, we sample one clip per video to learn a generalizable hypernetwork. For our models, we further sample one patch tubelet per clip.
When making a prediction for the weights of the HypoNeRV, each clip is treated independently, which allows for encoding of videos of arbitrary length. 

\subsection{General Configuration Settings}

Both NeRV-Enc* and our models use the same hypernetwork configuration, as defined in~\cite{chen2024fast}.
Specifically, the Transformer encoder consists of 6 blocks and 12 heads, a token dimension of 720, and feedforward layers of dimension 2880, with $32{\times}32$ patch tokenization.
Depending on the size of the learnable base and unique parameters, the hypernetwork size ranges from 41.1M to 47.6M parameters in our configurations.
The design of the hyponetwork, including number of NeRV blocks, upsampling strides, and token distribution, varies depending on target resolution and whether the model is patch-based (ours) or not (baseline).
All HypoNeRV configurations use GELU~\cite{hendrycks2016gaussian} activation.
For fair comparison, we carefully calibrate parameter distributions for NeRV-Enc* and TeCoNeRV models at each resolution to approximately match pre-quantization bpp.

Below, we provide comprehensive configuration details for models evaluated in \Cref{tab:allres_nonoverlap,tab:train_time_mem,tab:detail_nervenc}.

\subsection{NeRV-Enc* Configuration}
All baseline HypoNeRV configurations use 5 NeRV blocks.

\noindent\textbf{480p Resolution.}
For the primary NeRV-Enc* configuration evaluated at 480p in \Cref{tab:allres_nonoverlap}, we use the following settings:
\begin{itemize}
    \item HypoNeRV position embedding dimension: 32, channel width ($fc_{\text{dim}}$): 32
    \item HypoNeRV kernel sizes: 1, 3, 3, 3, 3 (layers one to five) 
    \item upscale factors (PixelShuffle strides): 5, 4, 4, 3, 2 (height) and 5, 4, 4, 4, 2 (width), for video size of $640{\times}480$
    \item HypoNeRV layers unique token numbers: 32, 256, 32, 24, 0; token dimensions: 200, 288, 288, 288, 0 (first layer to last)
    \item Hypernetwork of size 43.0M, 436.8K base parameters, 96.3K unique parameters
    \item Batch size: 8
\end{itemize}

\noindent For the smaller NeRV-Enc* configuration shown in \Cref{fig:rd_plot_nervenc} at 480p, we use the following settings:
\begin{itemize}
    \item HypoNeRV position embedding dimension: 20, channel width ($fc_{\text{dim}}$): 20
    \item HypoNeRV kernel sizes: 1, 3, 3, 3, 3 (layers one to five) 
    \item upscale factors (PixelShuffle strides): 5, 4, 4, 3, 2 (height) and 5, 4, 4, 4, 2 (width).
    \item HypoNeRV layers token numbers: 20, 160, 20, 20, 0; token dimensions: 125, 120, 288, 180, 0
    \item Hypernetwork of size 42.3M, 172.0K base parameters, 31.1K unique parameters
    \item Batch size: 8
\end{itemize}

\noindent\textbf{720p Resolution.} For the NeRV-Enc* configuration tested at 720p in \Cref{tab:allres_nonoverlap}, we use the following settings:
\begin{itemize}
    \item HypoNeRV position embedding dimension: 56, channel width ($fc_{\text{dim}}$): 56
    \item HypoNeRV kernel sizes: 1, 3, 3, 3, 3 (layers one to five) 
    \item HypoNeRV upscale factors (PixelShuffle strides): 5, 4, 4, 3, 3 (height) and 5, 4, 4, 4, 4 (width), for video size of $1280{\times}720$.
    \item HypoNeRV unique token numbers: 56, 448, 112, 112, 0;  unique token dimensions: 350, 504, 224, 168, 0
    \item Hypernetwork of size 47.6M, 1.3M base parameters, 289.0K unique parameters
    \item Batch size: 8
\end{itemize}

As described in \Cref{sub:main_res}, the memory required for scaling NeRV-Enc* to 1080p is significantly larger, hence we omit this comparison and provide comparison to NeRV and HiNeRV instead at this resolution. 
Further, true high-resolution uncompressed video data is expensive to store and load, making such a training process difficult to execute. 
This highlights the advantage of our method, which allows training at lower resolutions and predicting at higher resolutions, as detailed in \Cref{sub:res_indep}.

\subsection{TeCoNeRV Configuration}
Our patch-based models use a more compact 4-layer HypoNeRV architecture.
We report three configurations corresponding to different patch sizes for inference at multiple resolutions (\Cref{tab:allres_nonoverlap}).%

\noindent\textbf{TeCoNeRV\textsuperscript{1}: 320${\times}$160 patches (480p inference).}
For the $320{\times}160$ TeCoNeRV\textsuperscript{1} model evaluated at 480p inference in \Cref{tab:allres_nonoverlap}, we use the following settings. This is the same for the larger patch-tubelet and TeCoNeRV models in \Cref{fig:rd_plot_nervenc}.
\begin{itemize}
    \item HypoNeRV position embedding dimension: 14, channel width ($fc_{\text{dim}}$): 14
    \item HypoNeRV kernel sizes: 1, 3, 3, 3 (layers one to four) 
    \item HypoNeRV Upscale factors: 5, 4, 4, 2 (height) and 5, 4, 4, 4 (width)
    \item HypoNeRV unique token numbers: 5, 56, 4, 0; unique token dimensions: 196, 252, 196, 0 (first layer to last)
    \item Hypernetwork of size 41.3M, 65.2K base parameters, 15.9K unique parameters
    \item For $\mathcal{L}_{\text{temp}}$: $\lambda_{\text{temp}}$ = 0.1, applied to modulated parameters $\theta$
    \item Batch size: 32
\end{itemize}

\noindent For the smaller Patch-Tubelet and TeCoNeRV models used in \Cref{fig:rd_plot_nervenc} (Pareto front) at 480p, we keep the same settings as above and modify the HypoNeRV layers token numbers and token dimensions to be 5, 16, 4, 0 and 140, 252, 98, 0 respectively.

\noindent\textbf{TeCoNeRV\textsuperscript{2}: 320${\times}$240 patches (720p inference).}
For the $320{\times}240$ TeCoNeRV\textsuperscript{2} model evaluated at 720p inference in \Cref{tab:allres_nonoverlap}, we use the following settings.%
\begin{itemize}
    \item HypoNeRV position embedding dimension: 16, channel width ($fc_{\text{dim}}$): 20
    \item HypoNeRV kernel sizes: 1, 3, 3, 3 (layers one to four)
    \item HypoNeRV upscale factors: 5, 4, 4, 3 (height) and 5, 4, 4, 4 (width) 
    \item HypoNeRV unique token numbers: 10, 80, 16, 0; unique token dimensions: 200, 240, 240, 0 (first layer to last). 
    \item Hypernetwork of size 41.5M, 131.0K base parameters, 25.0K unique parameters
    \item For $\mathcal{L}_{\text{temp}}$: $\lambda_{\text{temp}}$ = 0.1, applied to modulated parameters $\theta$
    \item Batch size: 32
\end{itemize}

\noindent\textbf{TeCoNeRV\textsuperscript{3}: 384${\times}$270 patches (1080p inference).}
For the $384{\times}270$ TeCoNeRV\textsuperscript{3} model evaluated at 1080p inference in \Cref{tab:allres_nonoverlap}, we use the following settings.
\begin{itemize}
    \item HypoNeRV position embedding dimension: 20, channel width ($fc_{\text{dim}}$): 20
    \item HypoNeRV kernel sizes: 1, 3, 3, 3 (layers one to four) 
    \item Upscale factors: 6, 5, 3, 3 (height) and 6, 4, 4, 4 (width)
    \item HypoNeRV unique token numbers: 16, 100, 16, 0; unique token dimensions: 180, 240, 180, 0 (first layer to last)
    \item Hypernetwork of size 41.6M, 137.5K base parameters, 29.8K unique parameters
    \item For $\mathcal{L}_{\text{temp}}$: $\lambda_{\text{temp}}$ = 0.1, applied to modulated parameters $\theta$
    \item Batch size: 32
\end{itemize}

For additional implementation details, please refer to our code.

\section{Additional Results and Visualizations}

\subsection{Overlapping Strategies}
As discussed in \Cref{sub:quality_vs_size}, our patch-tubelet approach enables strided inference with overlap to eliminate boundary artifacts. While we adopt cropping as our primary fusion strategy, we also evaluate blending as an alternative.
Blending applies weighted averaging to overlapping regions, with position-dependent weights that fade smoothly toward patch edges.
This produces slightly smoother transitions between patches compared to cropping.

\Cref{tab:supp_overlap_patches} presents a detailed comparison of cropping versus blending across different overlap configurations at 720p and 1080p inference. 
We vary the overlap amount (in pixels, for height and width dimensions) and patch size to demonstrate the trade-off between visual quality and bitrate. 
Both strategies effectively remove boundary artifacts present in non-overlapping reconstructions. 
At 720p, blending provides marginal PSNR improvements (0.04-0.18dB) over cropping, while at 1080p, gains range from 0.02-0.18dB.
Encoded bitstream size (bpp) remains identical between both strategies.
However, cropping consistently achieves comparable or higher decoding FPS.
We therefore adopt cropping as our default strategy for reconstruction with overlap.
Encoding/decoding speeds are measured on an NVIDIA RTX A5000 for 720p inference, and RTX A6000 for 1080p inference.

\begin{table}
\centering
\caption{
\textbf{Overlapping strategies on UVG.} Varying overlap amount in pixels (height, width) and number of patches for overlap inference at 720p and 1080p. We compare cropping and blending fusion strategies in terms of reconstruction quality (PSNR/SSIM), compressed size (bpp), and encoding/decoding speeds (FPS). 
Both strategies effectively remove boundary artifacts, with cropping offering better speed-quality trade-offs.
}
\label{tab:supp_overlap_patches}
\resizebox{0.85\linewidth}{!}{
\begin{tabular}{@{}l cc cc cc@{}}
\toprule
& & & \multicolumn{2}{c}{Cropping} & \multicolumn{2}{c}{Blending} \\
\cmidrule(l){4-5}
\cmidrule(l){6-7}

\makecell{Overlap \\ (h, w)} & \#Patches & bpp $\downarrow$ & PSNR / SSIM $\uparrow$ & FPS $\uparrow$ & PSNR / SSIM $\uparrow$ & FPS $\uparrow$ \\
\midrule
\rowcolor{lightgray!30}  \multicolumn{7}{c}{\textit{Inference at 720p} (Patch Size: $320{\times}160$)} \\ 
\midrule

5, 0 & 20 & 0.0787 & 26.38 / 0.7489 & 51.8 / 171.4 & 26.42 / 0.7509 & 42.3 / 128.6 \\
20, 0 & 20 & 0.0788 & 26.39 / 0.7500 & 50.4 / 170.0 & 26.44 / 0.7526 & 53.6 / 129.2 \\
0, 5 & 25 & 0.0985 & 26.37 / 0.7467 & 42.6 / 118.4 & 26.54 / 0.7530 & 43.1 / 86.0 \\
5, 5 & 25 & 0.0985 & 26.42 / 0.7508 & 41.1 / 140.4 & 26.59 / 0.7575 & 34.7 / 86.1 \\
10, 10 & 25 & 0.0984 & 26.44 / 0.7523 & 35.8 / 113.8 & 26.61 / 0.7594 & 34.6 / 83.2 \\
20, 20 & 25 & 0.0985 & 26.45 / 0.7533 & 34.1 / 114.0 & 26.63 / 0.7608 & 32.4 / 83.5 \\

\midrule
\rowcolor{lightgray!30}  \multicolumn{7}{c}{\textit{Inference at 1080p} (Patch Size: $320{\times}240$)} \\ 
\midrule

5, 0 & 30 & 0.0728 & 26.54 / 0.7498 & 26.8 / 91.6 & 26.57 / 0.7517 & 27.5 / 67.9 \\
20, 0 & 30 & 0.0728 & 26.84 / 0.7542 & 20.8 / 74.9 & 26.86 / 0.7566 & 27.1 / 50.7 \\
5, 5 & 35 & 0.0849 & 26.57 / 0.7515 & 22.3 / 77.0 & 26.69 / 0.7569 & 22.2 / 49.9 \\
10, 10 & 35 & 0.0850 & 26.67 / 0.7537 & 22.4 / 74.6 & 26.85 / 0.7598 & 22.5 / 52.5 \\
20, 20 & 35 & 0.0850 & 26.88 / 0.7572 & 22.3 / 75.8 & 27.01 / 0.7631& 22.6 / 53.7 \\

\bottomrule
\end{tabular}
}
\end{table}

\subsection{Additional Qualitative Results}

\begin{figure}
\centering

    \begin{subfigure}{\linewidth}
        \centering
        \includegraphics[width=\linewidth]{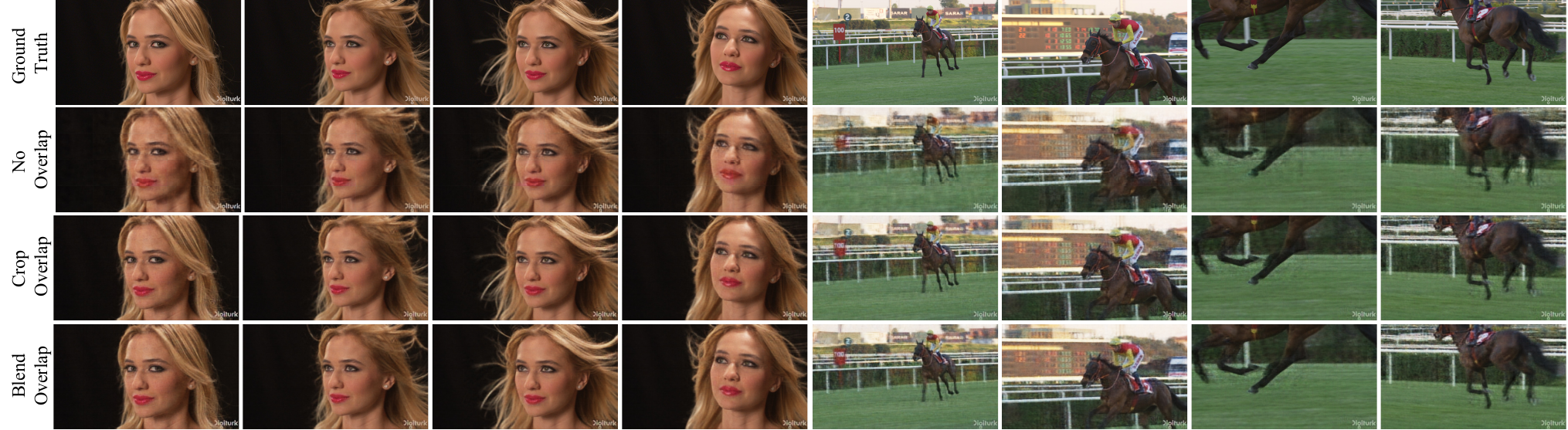}
        \subcaption{Beauty (left), Jockey (right) sequences from UVG.}
    \end{subfigure}
    
    \vspace{0.5em} 

    \begin{subfigure}{\linewidth}
        \centering
        \includegraphics[width=\linewidth]{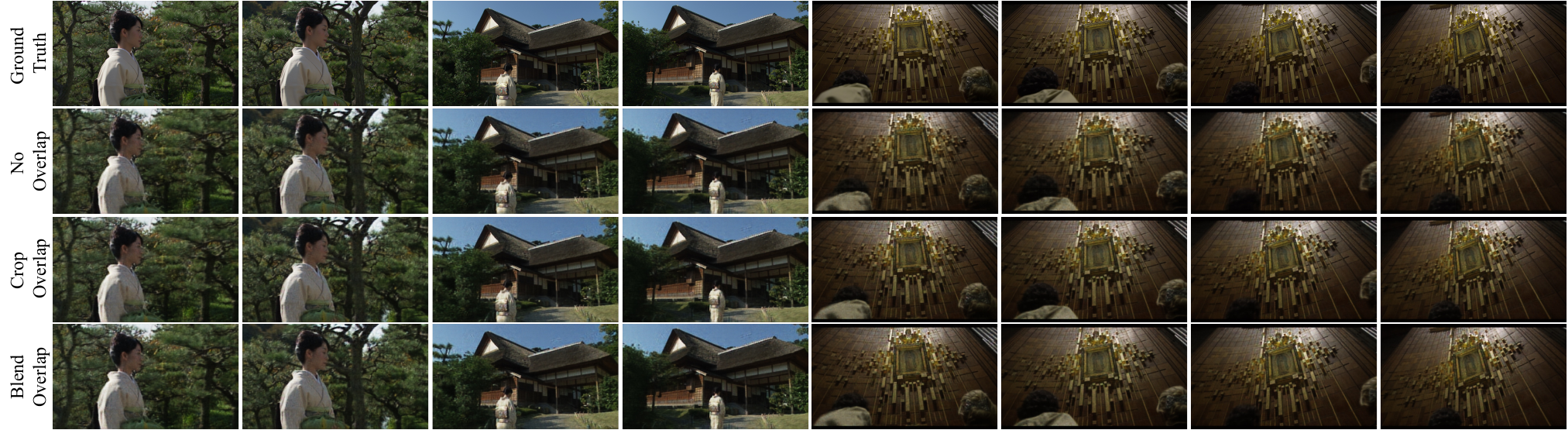}
        \subcaption{Kimono sequence from HEVC (left), videoSRC03 sequence from MCL-JCV (right).}
    \end{subfigure}

    \caption{\textbf{Qualitative results for TeCoNeRV at 1080p.} 
    Comparison of non-overlapping inference and overlapping strategies (crop and blend) for TeCoNeRV on sequences from (a) UVG and (b) HEVC, MCL-JCV. 
    Top to bottom for each sequence: Ground truth, ours (“no overlap”), ours (“overlap with cropping”), ours (“overlap with blending”). 
    Overlapping effectively eliminates boundary artifacts that may be visible in non-overlapping predictions, with blending producing slightly smoother transitions than cropping.
    }
    \label{fig:supp_qual_1080p} 

\end{figure}

\Cref{fig:supp_qual_1080p} shows qualitative results for TeCoNeRV at 1080p across sequences from UVG, HEVC and MCL-JCV, demonstrating that overlapping strategies -- cropping or blending -- successfully eliminate the few boundary artifacts visible in non-overlapping inference.

\section{Ablations}
\label{sec:supp_ablation}

\subsection{Patch Tubelet Size}

\begin{table}
\centering
\caption{\textbf{Effect of patch tubelet dimensions on reconstruction quality and compression on UVG at 480p}. 
We evaluate different spatial partitioning strategies with varying base and unique parameter allocations.
Our chosen $320{\times}160$ configuration optimally balances reconstruction quality, compression efficiency, and total patch count relative to frame resolution.
}
\label{tab:supp_ablate_patch}
\resizebox{0.75\linewidth}{!}{
\begin{tabular}{@{}c cc c c c c @{}}
\toprule
{Tubelet size} & {PE, Ch} & {\#Base} & {\#Unique} & {PSNR / SSIM $\uparrow$} & {bpp $\downarrow$} \\
\midrule
$320{\times}160$ & 14, 14 & 65.2K  & 15.87K & 25.49 / 0.7159 & 0.0826 \\
$160{\times}160$ & 14, 14 & 63.7K  & 15.87K &  26.06 / 0.7310 & 0.1708  \\
$160{\times}160$ & 14, 14 & 63.7K  & 7.94K  &  25.15 / 0.6981 & 0.0944  \\
$320{\times}240$ & 16, 20 & 130.9K & 22.0K & 24.17 / 0.6836 & 0.0737 \\

$160{\times}160$ & 8, 8 & 21.4K & 10.0K & 24.52 / 0.6780 & 0.1258 \\
$160{\times}160$ & 8, 8 & 21.4K & 5.39K & 24.11 / 0.6671 & 0.0724 \\
$160{\times}120$ & 8, 8 & 19.0K & 6.56K & 21.88 / 0.6132 & 0.1253 \\
$160{\times}120$ & 14, 14 & 56.6K & 8.43K & 22.21 / 0.6294 & 0.1517 \\
$160{\times}80$ & 12, 12 & 36.6K & 6.36K & 22.41 / 0.6282 & 0.1842 \\

$80{\times}80$ & 8, 8 & 14.4K & 1.93K & 21.64 / 0.6088 & 0.1196 \\
$80{\times}80$ & 4, 4 & 3.9K & 1.54K & 21.00 / 0.5724 & 0.0978 \\
\bottomrule
\end{tabular}
}
\end{table}

\Cref{tab:supp_ablate_patch} examines different patch-tubelet dimensions for encoding the UVG dataset at 480p, varying both spatial partitioning strategies and parameter allocations. 
These experiments were conducted using our patch-tubelet model without temporal coherence finetuning.
Smaller patches achieve higher PSNR but at increased bitrate due to the greater number of patches required. 
Larger patches reduce bitrate but may struggle to capture sufficient spatial detail for high-quality reconstruction, even with increased parameter counts. 
Our chosen $320{\times}160$ configuration optimally balances reconstruction quality, compression efficiency, and practical considerations including memory requirements and total number of patches relative to target resolution.
We observe similar trends when conducting these ablations at 720p and 1080p resolutions, which we omit for brevity.

\subsection{Clip Length}
Table~\ref{tab:suppl_ablate_frame_num} explores varying number of frames in a clip -- 4, 12, and 16 frames compared to our main 8-frame configuration. 
For each clip length, we test multiple base and unique parameter configurations to ensure fair comparison across different bitrates. 
Shorter clips can achieve higher reconstruction quality but at increased bitrates.
Longer clips reduce bitrates but may diminish quality.
While the 4-frame model with reduced unique parameters (row 5: 25.79dB, 0.0856 bpp) marginally outperforms our chosen configuration, it possesses a high base-to-unique parameter ratio (65.2K vs 7.94K), also requiring twice as many predictions per video.

\begin{table}
\centering
\caption{\textbf{Impact of clip length on reconstruction quality and compression on UVG at 480p.}
We vary the number of consecutive frames per clip with different parameter configurations. 
Our 8-frame configuration balances quality, compression efficiency, and computational overhead. 
PE and Ch denote positional embedding dimension and convolution channel width.
}
\label{tab:suppl_ablate_frame_num}
\resizebox{0.7\linewidth}{!}{
\begin{tabular}{@{}c cc c c c c @{}}
\toprule
{\#Frames} & {PE, Ch} & {\#Base} & {\#Unique} & {PSNR/SSIM $\uparrow$} & {bpp $\downarrow$} \\
\midrule
8 & 14, 14 & 65.2K  & 15.87K & 25.49 / 0.7159 & 0.0826 \\
4 & 14, 14 & 65.2K  & 15.87K & 26.75 / 0.7457 & 0.1523 \\
12 & 14, 14 & 65.2K & 15.87K  & 24.18 / 0.6662 & 0.0606 \\
16 & 14, 14 & 65.2K & 15.87K  & 23.13 / 0.6517 & 0.0464 \\
4 & 14, 14 & 65.2K & 7.94K & 25.79 / 0.7187 & 0.0856 \\
4 & 8, 8 & 22.2K & 5.92K & 24.90 / 0.6940 & 0.0702 \\

12 & 14, 16 & 83.7K & 23.92K & 24.22 / 0.6682 & 0.0909 \\
12 & 16, 18 & 105.5K & 21.24K & 24.71 / 0.6877 & 0.0776 \\
12 & 18, 18 & 106.3K & 20.49K & 24.94 / 0.6945 & 0.0726 \\

16 & 14, 14 & 65.2K & 32.24K & 23.00 / 0.6494 & 0.0935 \\
16 & 16, 20 & 128.7K & 28.0K & 24.95 / 0.7067 & 0.0757 \\

\bottomrule
\end{tabular}
}
\end{table}

\subsection{Network Parameters}

Table~\ref{tab:ablate_net_param_size} investigates how parameter allocation influences compression performance on UVG at 480p. 
We present variations both in base parameters (controlled by time positional embedding dimension and convolution channel width) and unique parameters (determined by token number and dimensions distributed across hyponetwork layers). 
For all configurations, we use $320{\times}160$ patches and a 4-layer HypoNeRV architecture with horizontal upsampling strides of 5, 4, 4, 4 and vertical upsampling strides of 5, 4, 4, 2.
All experiments are conducted without temporal coherence finetuning.

Our selected configuration (first row: 65.2K base, 15.87K unique) provides a desirable quality-compression balance with minimal base parameter overhead. 
Increasing base parameters can improve PSNR, but with diminishing returns as parameter count grows.
However, inflating base parameters while constraining unique parameters to match target bitrates is suboptimal, as larger base parameters increase memory consumption and computational overhead during both training and inference.
Notably, even our most lightweight configuration (seventh row: 5.12K unique parameters) achieves 23.21dB at 0.0335 bpp, maintaining quality comparable to NeRV-Enc* while reducing bitrate by nearly 70\%. 
These results demonstrate that strategic parameter allocation is important for efficient implicit neural video compression~\cite{kim2022generalizable}.

\begin{table}
\centering
\caption{\textbf{Ablation on network parameter configurations for 480p encoding on UVG.} 
We evaluate the impact of different parameter size allocations on the quality-compression tradeoff. 
We vary positional embedding dimensions (PE), channel widths (Ch), and the distribution of unique parameters across layers. 
All models use a 4-layer HypoNeRV architecture and $320{\times}160$ patches. 
Our selected configuration (first row) achieves strong quality-compression trade-offs with minimal base parameter overhead.
}
\label{tab:ablate_net_param_size}
\resizebox{0.85\linewidth}{!}{
\begin{tabular}{@{}cc c cc c c@{}}
\toprule
 &  &  & \multicolumn{2}{c}{\# Total Params} & \multicolumn{2}{c}{Metrics}  \\
\cmidrule(l){4-5}
\cmidrule(l){6-7}
 PE & Ch &  Layerwise Unique Params & Base & Unique & PSNR / SSIM $\uparrow$ & bpp $\downarrow$ \\
\midrule
14 & 14 & 0.98K, 14.1K, 0.78K, 0  & 65.2K & 15.87K & 25.49 / 0.7159 & 0.0826 \\ 

14 & 14 & 0.98K, 14.1K, 0.78K, 0.75K  & 65.2K & 16.63K & 24.81 / 0.6953 & 0.0938 \\ 

14 & 14 & 0.98K, 14.1K, 0.78K, 3.02K  & 65.2K & 18.90K & 24.73 / 0.6886 & 0.1103 \\ 

16 & 16 & 1.28K, 12.29K, 3.07K, 0 & 128.7K & 16.64K & 26.22 / 0.7353 & 0.0900 \\ 

16 & 20 & 0.8K, 14.40K, 0.8K, 0 & 128.7K & 16.0K & 25.16 / 0.7040 & 0.0809 \\ 
16 & 20 & 0.8K, 19.2K, 2.3K, 0 & 128.7K & 22.3K & 25.36 / 0.7067 & 0.1163 \\ 

14 & 14 & 0.7K, 4.03K, 0.39K, 0 & 65.2K & 5.12K & 23.21 / 0.6443 & 0.0335 \\ 
8 & 8 & 0.8K, 4.61K, 0.51K, 0 & 22.2K & 5.92K & 23.18 / 0.6482 & 0.0395 \\ 

\bottomrule
\end{tabular}
}
\end{table}

\subsection{Temporal Coherence Loss Components}

\begin{table}
\centering
\caption{\textbf{Ablation on distance metrics for temporal coherence regularization}. 
We evaluate different weightings of $\ell_1$ (absolute differences) and $\ell_2$ (squared differences) norms in our temporal coherence loss on the $320{\times}160$ TeCoNeRV configuration for inference on UVG at $480$p. 
Our chosen configuration ($\lambda_{\ell_1}=0.1,\lambda_{\ell_2}=0$) provides optimal rate-distortion performance, with higher values providing smaller size at the expense of quality.
Our default configuration uses only $\ell_1$ loss for the sake of simplicity, inducing sparsity in residuals.
}
\label{tab:suppl_ablate_l1_l2}
\small
\resizebox{0.48\linewidth}{!}{
\begin{tabular}{@{}cc ccc @{}}
\toprule
$\lambda_{\ell_1}$ & $\lambda_{\ell_2}$ & PSNR $\uparrow$ & SSIM $\uparrow$ & bpp $\downarrow$ \\
\midrule
0.07  & 0   & 25.62 & 0.7133 & 0.0680  \\
0.07  & 0.1 & 25.60 & 0.7127 & 0.0683  \\
0.08  & 0   & 25.57 & 0.7119 & 0.0681  \\
0.08  & 0.1 & 25.56 & 0.7118 & 0.0683  \\
0.09  & 0.1	& 25.53	& 0.7109 & 0.0680  \\
0.1   & 0.0 & 25.52 & 0.7103 & 0.0676  \\
0.1   & 0.1 & 25.51 & 0.7102 & 0.0676  \\
0.12  & 0   & 25.46 & 0.7087 & 0.0662  \\
0.2   & 0   & 25.30 & 0.7039 & 0.0631  \\
0.2   & 0.2 & 25.29 & 0.7036 & 0.0633  \\
0.3   & 0   & 25.13 & 0.6988 & 0.0632  \\
1.0   & 0   & 24.53 & 0.6812 & 0.0581  \\
2.0   & 0   & 24.07 & 0.6684 & 0.0552  \\
3.0   & 0   & 23.72 & 0.6596 & 0.0490  \\
5.0   & 0   & 22.96 & 0.6411 & 0.0474  \\

\bottomrule
\end{tabular}
}
\end{table}

\begin{table}
\centering
\caption{\textbf{Impact of regularization target on compression performance}. 
We compare applying temporal coherence regularization to unique parameters ($\theta_{\text{uniq}}$), modulated parameters ($\theta$), or both, across different quantization levels. 
We ablate on UVG at both 480p and 720p resolutions.
Regularizing modulated parameters yields the best compression efficiency at lower bit depths.
}
\label{tab:supp_ablate_reg_mode}
\resizebox{0.75\linewidth}{!}{
\begin{tabular}{@{}r cc cc @{}}
\toprule
 & \multicolumn{2}{c}{8 bit quant} & \multicolumn{2}{c}{4 bit quant} \\
 \cmidrule(l){2-3}
 \cmidrule(l){4-5}
 \makecell{Regularization\\Parameter Type} & PSNR/SSIM $\uparrow$ & bpp $\downarrow$ & PSNR/SSIM $\uparrow$ & bpp $\downarrow$ \\
\midrule
\rowcolor{lightgray!30}  \multicolumn{5}{c}{\textit{480p}} \\
\midrule
$\theta$  &  
26.18 / 0.7318 & 0.2005 &  
25.52 / 0.7103 & 0.0676 \\

$\theta_{\text{uniq}}$  &  
26.20 / 0.7328 & 0.2001 & 
25.79 / 0.7202 & 0.0718 \\

$\theta_{\text{uniq}}$ + $\theta$ & 
26.11 / 0.7288 & 0.1911 & 
25.52 / 0.7095 & 0.0671 \\

\midrule
\rowcolor{lightgray!30}  \multicolumn{5}{c}{\textit{720p}} \\
\midrule

$\theta$   &  
25.68 / 0.7364 & 0.2035 &  
25.22 / 0.7242 & 0.0667 \\

$\theta_{\text{uniq}}$  &
25.69 / 0.7400 & 0.2031 &
25.38 / 0.7324 & 0.0716 \\

$\theta_{\text{uniq}}$ + $\theta$ &
25.52 / 0.7335 & 0.1914 &
25.08 / 0.7220 & 0.0663 \\

\bottomrule
\end{tabular}
}
\end{table}

\noindent\textbf{Regularization strength.}
As described in \Cref{sec:temp_coh}, our temporal coherence loss (\Cref{eqn:temp_finetune}) is controlled by the hyperparameter $\lambda_{\text{temp}}$, which balances reconstruction quality against smoothness of weight transitions. 
This parameter provides a mechanism for rate control: higher values of $\lambda_{\text{temp}}$ enforce stronger temporal coherence, producing smaller residuals between consecutive clips and thus lower bitrates, at the cost of some reconstruction quality.
Through hyperparameter search, we determine that $\lambda_{\text{temp}} = 0.1$ provides optimal quality-compression trade-offs for our main results. 
However, $\lambda_{\text{temp}}$ can be adjusted based on requirements. \Cref{fig:l1_lever_overlap_combined} demonstrates this trade-off on overlapping inference configurations. Table~\ref{tab:suppl_ablate_l1_l2} extends this analysis with a broader range of $\lambda_{\text{temp}}$ values up to 5.0 (no overlap, UVG at 480p), demonstrating the full spectrum of quality-bitrate trade-offs achievable through temporal coherence regularization.

\noindent\textbf{Choice of distance metric.}
While our main settings for $\mathcal{L}_\text{temp}$ use the $\ell_1$ norm for measuring weight differences (\Cref{eqn:temp_loss}), Table~\ref{tab:suppl_ablate_l1_l2} also evaluates the $\ell_2$ norm and their combination. The $\ell_1$ norm induces sparsity by encouraging smaller weight differences to become exactly zero, resulting in more compressible residuals. 
The $\ell_2$ norm penalizes large differences more heavily but encourages more evenly distributed magnitudes across all weights. 
While $\ell_2$ regularization can provide marginal gains in certain configurations, $\ell_1$ consistently produces more compact bitstreams. 
We adopt $\ell_1$ as our primary distance metric for its superior compression efficiency and simplicity.

\noindent\textbf{Regularization parameter mode.} 
\Cref{tab:supp_ablate_reg_mode} investigates applying regularization to different parameter sets: pre-modulation parameters $\theta_{\text{uniq}}$, modulated parameters $\theta$, or both.
We evaluate on UVG at both 480p resolution ($320{\times}160$ patches) and 720p resolution ($320{\times}240$ patches).
Regularizing modulated parameters yields the best compression efficiency, particularly at lower bit depths, while regularizing both produces similar results. 
For simplicity, we regularize the modulated parameters in our final model.

\begin{table}
\centering
\caption{\textbf{Entropy coding performance at varying quantization levels, on UVG at 480p.} Arithmetic coding obtains marginally better bpp compared to Huffman coding across all bit depths, with the advantage increasing at lower quantization levels. 
Results demonstrate the complementary relationship between our temporal coherence approach and advanced entropy coding techniques.}
\label{tab:ablate_huffman_quant}
\resizebox{0.7\linewidth}{!}{
\begin{tabular}{@{}c cc cc cc@{}}
\toprule
Quant Level & PSNR/SSIM $\uparrow$ & bpp (Arithmetic) $\downarrow$ & bpp (Huffman) $\downarrow$ \\
\midrule
8 & 26.18 / 0.7318 & 0.2005 & 0.2017 \\
7 & 26.17 / 0.7314 & 0.1633 & 0.1646 \\
6 & 26.13 / 0.7301 & 0.1275 & 0.1290 \\
5 & 25.98 / 0.7253 & 0.0948 & 0.0969 \\
4 & 25.52 / 0.7103 & 0.0676 & 0.0718 \\
\bottomrule
\end{tabular}
}
\end{table}

\begin{table}
\small
\centering
\caption{\textbf{Results on individual videos of the UVG dataset at 480p resolution.} 
TeCoNeRV consistently outperforms the baseline (NeRV-Enc*) across all videos while reducing bitrates.}
\label{tab:supp_per_vid_uvg}
\resizebox{\linewidth}{!}{
\begin{tabular}{@{}l cc cc cc@{}}
\toprule
& \multicolumn{2}{c}{NeRV-Enc*~\cite{chen2024fast}} & \multicolumn{2}{c}{Ours (Patch-Tubelet)} & \multicolumn{2}{c}{Ours (TeCoNeRV)} \\
\cmidrule(lr){2-3} \cmidrule(lr){4-5} \cmidrule(lr){6-7}
Video & PSNR/SSIM $\uparrow$ & bpp $\downarrow$ & PSNR/SSIM $\uparrow$ & bpp $\downarrow$ & PSNR/SSIM $\uparrow$ & bpp $\downarrow$ \\
\midrule 
Beauty & 27.52 / 0.7822 & 0.1050 & 31.06 / 0.8268 & 0.0740 & 30.64 / 0.8138 & 0.0596 \\
Bosphorus & 25.59 / 0.7120 & 0.1030 & 28.26 / 0.7798 & 0.0734 & 28.14 / 0.7678 & 0.0620 \\
HoneyBee & 21.32 / 0.5932 & 0.1021 & 24.25 / 0.7102 & 0.0797 & 24.33 / 0.7250 & 0.0679 \\
Jockey & 21.78 / 0.7216 & 0.1106 & 23.68 / 0.7504 & 0.0949 & 23.91 / 0.7336 & 0.0739 \\
ShakeNDry & 23.81 / 0.5252 & 0.1025 & 26.43 / 0.6391 & 0.0791 & 26.22 / 0.6291 & 0.0680 \\
YachtRide & 23.28 / 0.6363 & 0.1047 & 25.25 / 0.6941 & 0.0808 & 25.28 / 0.6857 & 0.0673 \\
ReadySteadyGo & 18.43 / 0.5125 & 0.1096 & 20.00 / 0.5732 & 0.0947 & 20.46 / 0.5767 & 0.0746 \\
\bottomrule
\end{tabular}
}
\end{table}

\subsection{Entropy Coding Strategies}
\label{supp_subsec:huffman}
We evaluate the impact of two different entropy coding techniques on the final compression efficiency of our models. 
Table~\ref{tab:ablate_huffman_quant} shows the performance of our $320{\times}160$ TeCoNeRV configuration across different quantization levels using both of these coding methods, on UVG at 480p. 
Arithmetic coding yields better compression (slightly lower bpp) than Huffman coding, with the advantage becoming more pronounced at lower bit depths, while maintaining identical reconstruction quality (as both methods are lossless transformations of the same quantized data).

\subsection{Results on Individual UVG Videos}

In \Cref{tab:supp_per_vid_uvg}, we present the PSNR, SSIM, and bits-per-pixel (bpp) metrics for each of the 7 videos in UVG at 480p, comparing our methods against the NeRV-Enc* baseline.
Both our Patch-Tubelet model and TeCoNeRV obtain better reconstruction quality-size tradeoff compared to the baseline across all sequences.
TeCoNeRV further reduces bitrates (by 14-22\% across videos) using temporal coherence regularization, with minimal impact on reconstruction quality.

\end{document}